\documentclass{article}

\usepackage{microtype}
\usepackage{graphicx}
\usepackage{subfigure}
\usepackage{booktabs} 

\usepackage{hyperref}

\usepackage[preprint]{icml2026}

\usepackage{amsmath}
\usepackage{amssymb}
\usepackage{mathtools}
\usepackage{amsthm}
\usepackage{amsmath}
\usepackage{graphicx}
\usepackage{wrapfig}
\usepackage{amsmath, bm}

\usepackage{listings}
\usepackage{xcolor}
\usepackage{booktabs} 

\usepackage[capitalize,noabbrev]{cleveref}

\theoremstyle{plain}

\theoremstyle{definition}

\theoremstyle{remark}

\usepackage[textsize=tiny]{todonotes}

\definecolor{codegreen}{rgb}{0,0.6,0}
\definecolor{codegray}{rgb}{0.5,0.5,0.5}
\definecolor{codepurple}{rgb}{0.58,0,0.82}
\definecolor{backcolour}{rgb}{0.95,0.95,0.92}

\begin{document}

\twocolumn[
\icmltitle{Online Vector Quantized Attention}



  \icmlsetsymbol{equal}{*}

  \begin{icmlauthorlist}
    \icmlauthor{Nick Alonso}{z}
    \icmlauthor{Tomas Figliolia}{z}
    \icmlauthor{Beren Millidge}{z}
  \end{icmlauthorlist}

  \icmlaffiliation{z}{Zyphra}
  
  \icmlcorrespondingauthor{Nick Alonso}{nick@zyphra.com}

  \icmlkeywords{linear attention, attention, language model, llm}

  \vskip 0.3in
]



\printAffiliationsAndNotice{}




\begin{abstract}
Standard sequence mixing layers used in language models struggle to balance efficiency and performance. Self-attention performs well on long context tasks but has expensive quadratic compute and linear memory costs, while linear attention and SSMs use only linear compute and constant memory but struggle with long context processing. In this paper, we develop a sequence mixing layer that aims to find a better compromise between memory-compute costs and long-context processing, which we call online vector-quantized (OVQ) attention. OVQ-attention requires linear compute costs and constant memory, but, unlike linear attention and SSMs, it uses a sparse memory update that allows it to greatly increase the size of its memory state and, consequently, memory capacity. We develop a theoretical basis for OVQ-attention based on Gaussian mixture regression, and we test it on a variety of synthetic long context tasks and on long context language modeling. OVQ-attention shows significant improvements over linear attention baselines and the original VQ-attention, on which OVQ-attention was inspired. It demonstrates competitive, and sometimes identical, performance to strong self-attention baselines up 64k sequence length, despite using a small fraction of the memory of full self-attention. 
\end{abstract}

\section{Introduction}
Sequence mixing layers allow LLMs to utilize information across input tokens to perform complex tasks. Predominant sequence mixing layers used in LLMs currently struggle to balance computational efficiency and performance. Self-attention \cite{bahdanau2014neural, vaswani2017attention} shows an impressive ability to utilize long sequences for tasks requiring in-context learning \cite{dong2024survey}, reasoning \cite{chen2025towards}, and long-range information retrieval \cite{hsieh2024ruler}. However, self-attention incurs an expensive quadratic compute cost and linearly growing memory. Linear attention \cite{katharopoulos2020transformers, shen2021efficient, yang2024parallelizing, yang2024gated, von2025mesanet} and SSM models \cite{gu2024mamba, dao2024transformers}, on the other hand, are highly efficient, incurring only linear compute and a relatively small constant memory cost. However, the small capacity of these layers leads to large performance deficits on tasks requiring precise and accurate processing of information over long contexts (e.g., \cite{akyurekcontextlang, wang2025systematic}). Hybrid models \cite{glorioso2024zamba, lieber2024jamba,glorioso2024zamba2, team2025kimi}, that interleave self-attention and SSM layers, alleviate, but do not eliminate, the quadratic compute and linear memory costs of the self-attention layers.

In this paper, we develop a novel approach to sequence mixing that aims to find a better compromise between memory/compute costs and long-context processing ability. Our approach builds upon a little-explored sequence mixing layer called vector quantized attention \cite{lingleVQ} (VQ-attention), which, like linear attention and SSM models, has constant memory and linear compute costs. However, unlike these methods, VQ-attention uses a sparse memory update, based on k-means clustering. The state update is decoupled from memory size,allowing for the use of very large memory states and, potentially, storage capacity while being computationally tractable and efficient. VQ-attention stores a dictionary of centroids modeling keys and values. The key dictionary, $D_k$, is learned during pretraining, while the value dictionary, $D_v$, is learned online during the forward pass. 

Despite this potential, we show the original form of VQ-attention struggles on simple long-context processing tasks. We find the main issue is related to the static, pretrained key dictionary $D_k$, which struggles to model keys well during the forward pass. Motivated by this, we develop an alternative form of VQ-attention, which we call online VQ-attention (OVQ-attention) which learns both the key and value dictionaries online. We demonstrate that OVQ attention substantially outperforms original VQ attention and can be competitive with full attention at long context lenghts.

Our contributions in this paper can be described as follows:

\begin{enumerate}
\item We propose an alternative form of VQ-attention, called online VQ-attention (OVQ-attention), which learns both $D_k$ and $D_v$ online during the forward pass. We develop a novel theoretical framework for OVQ-attention based on Gaussian Mixture Regression and use this theoretical basis to develop a principled online learning algorithm for the OVQ-attention layer, involving novel methods for growing the memory state $[D_k, D_v]$ asymptotically toward a constant maximum size and for updating the memory states using sparse online update rules.
\item We test our OVQ-attention layer's ability to perform long context processing, using a variety of synthetic tasks testing in-context recall (ICR) and long range in-context learning (ICL), as well as long-context language modeling. Where OVQ-attention shows significant improvements in performance over VQ-attention and baseline linear attention models, across all tasks. OVQ-attention matches or slightly deviates from strong baselines that use full self-attention, even at long test lengths when only using a state size 10-25$\%$ of that used by self-attention. 
\item OVQ-attention also demonstrates the ability to effectively length extrapolate from 4k tokens on ICR and ICL up to and beyond 64k tokens. We also find the amount of memory OVQ-attention uses can be dynamically adjusted at test time, with larger memory allocations consistently leading to better performance. 
\end{enumerate}

\section{Background}

\subsection{Self Attention}
Let $X \in \mathbb{R}^{T\times D}$ be the layer input, where $T$ is sequence length and $D$ is the model dimension. Let $Q, K, V \in \mathbb{R}^{T\times d}$, where $d$ is the dimension of each head. These are computed $Q=f(W_QX)$, $K=f(W_VX)$, $V=W_VX$, where $f$ may be the identity function, layer norm, or other non-linearity. The causal self-attention operation for input at position $t$ is
\begin{equation}
O_t^{\texttt{att}} = \texttt{softmax}(\beta \mathbf{q}_t K_{0:t}^{\top})V_{0:t},\\
\end{equation}
where $\mathbf{q}_t \in \mathbb{R}^{1\times d}$ is the query at position $t$. This operation is parallelized through the use of an attention mask:
\begin{equation}
O^{\texttt{att}} = \texttt{softmax}(\beta Q K^{\top}+M)V
\end{equation}
where softmax is applied row-wise. 

\subsection{Vector Quantized Attention}
Let $D_k \in \mathbb{R}^{N\times d}$ be a dictionary of centroids, $[\boldsymbol{\mu}^{k\top}_0, \boldsymbol{\mu}^{k\top}_1,...,\boldsymbol{\mu}^{k\top}_N]^{\top}$, used to quantize $K$ in the self attention operation, where $N$ is the number of centroids. VQ-attention treats $D_k$ as a set of parameters optimized during pretraining \cite{lingleVQ}. During inference, VQ-attention applies vector quantization to keys by replacing each key, $k_t$, with its nearest centroid. Let the set of quantized keys be $\hat{K}$. VQ-attention can be expressed in a form that has quadratic time complexity:
\begin{align}\label{eq:vq_quad}
O_t^{\texttt{vq-att}} &= \texttt{softmax}(\beta \mathbf{q}_t \hat{K}_{0:t}^{\top})V_{0:t},\\
O^{\texttt{vq-att}} &= \texttt{softmax}(\beta Q \hat{K}^{\top}+M)V.
\end{align}
The quadratic version of VQ-attention is simply self-attention with vector-quantized keys. However, crucially, the quadratic form of VQ-attention has an equivalent linear time and constant memory version \cite{lingleVQ}. The intuition is that when we vector-quantize the keys, we do not need to compare queries to each of the previous keys. We only need to compare queries to each dictionary centroid and track the number of keys assigned to each centroid. These two quantities allow us to get back output equivalent to that of the quadratic VQ-attention version (for proof see \cite{lingleVQ}):
\begin{align}
O_t^{\texttt{vq-att}} &= \texttt{softmax}(\beta \mathbf{q}_t \hat{K}_{0:t}^{\top})V_{0:t}\\&= \texttt{softmax}(\beta \mathbf{q}_t D_k^{\top} + \text{log}(\mathbf{c}_t))D_v,\label{eq:vq_lin1}
\end{align}
where $\mathbf{c}_t \in \mathbb{R}^{1 \times N}$ is a count vector representing the number of keys up to position $t$ assigned to each of the $N$ centroids, and $D_v \in \mathbb{R}^{N\times d}$ is a dictionary of centroids, $[\boldsymbol{\mu}^{v\top}_0, \boldsymbol{\mu}^{v\top}_1,...,\boldsymbol{\mu}^{v\top}_N]^{\top}$, used to quantize values. $D_v$, unlike $D_k$, computed on-the-fly during inference. Specifically, each value $\mathbf{v}_t$ in $V$ uses the same centroid assignment as the keys. Each centroid $\mathbf{\mu}_n^v$ is set to an average of all previous values assigned to centroid $n$. We can see that the right hand side of equation 5 uses a constant amount of compute per query, since $\mathbf{q}_t$ is compared to a fixed number of centroids ($N$).

\textbf{Chunk-Parallel Form.} Without a causal mask, the right side of equation \ref{eq:vq_lin1} can be parallelized as follows
\begin{equation}
O^{\texttt{vq-att}} = \texttt{softmax}(Q D_k^{\top} + \text{log}(C)) D_v,
\end{equation}
where tensor $C \in \mathbb{R}^{T\times N}$ are the cumulative counts storing the number of key-values assigned to each row of $D_v$ at each point in the sequence. Introducing a causal mask while retaining linear time complexity requires a more complex equation that uses chunk-level recurrence \cite{lingleVQ}. The sequence is split into $C$ chunks of a fixed chunk length, $L$. For each chunk $c$, the value dictionary $D_{v(c)}$ is updated recurrently using values from the current and previous chunks. The VQ-attention output is then computed as follows 
\begin{align}
O^{\texttt{vq-att}} = \frac{1}{Z} &(\text{exp}(Q_c D_k^{\top} +  \text{log}(\mathbf{c}_{c-2})) D_{v(c-2)}\\
&+ \text{exp}(Q_c \hat{K}_{c-1}^{\top} + M_{c-1})V_{c-1}\\
&+ \text{exp}(Q_c \hat{K}_{c}^{\top} + M_{c})V_{c}),
\end{align}
where $\frac{1}{Z}$ is a vector that does the softmax normalization and $\mathbf{c}_{c-2}$ stores the counts up until the last key-value of chunk $c-2$. $M_{c-1}$ and $M_{c}$ create a causal sliding window mask such that future keys are masked and each query can attend to the previous $L$ quantized keys, represented in equation 5 and 6, and dictionary elements, shown in line 4. Each operation (line 7, 8, and 9) has linear time complexity, which entails this chunk-wise recurrent formulation of VQ-attention also has linear time complexity \cite{lingleVQ}.

\subsection{Preliminary Testing: In-Context Recall}
\begin{figure}[h]
  \begin{center}
\includegraphics[width=0.38\textwidth]{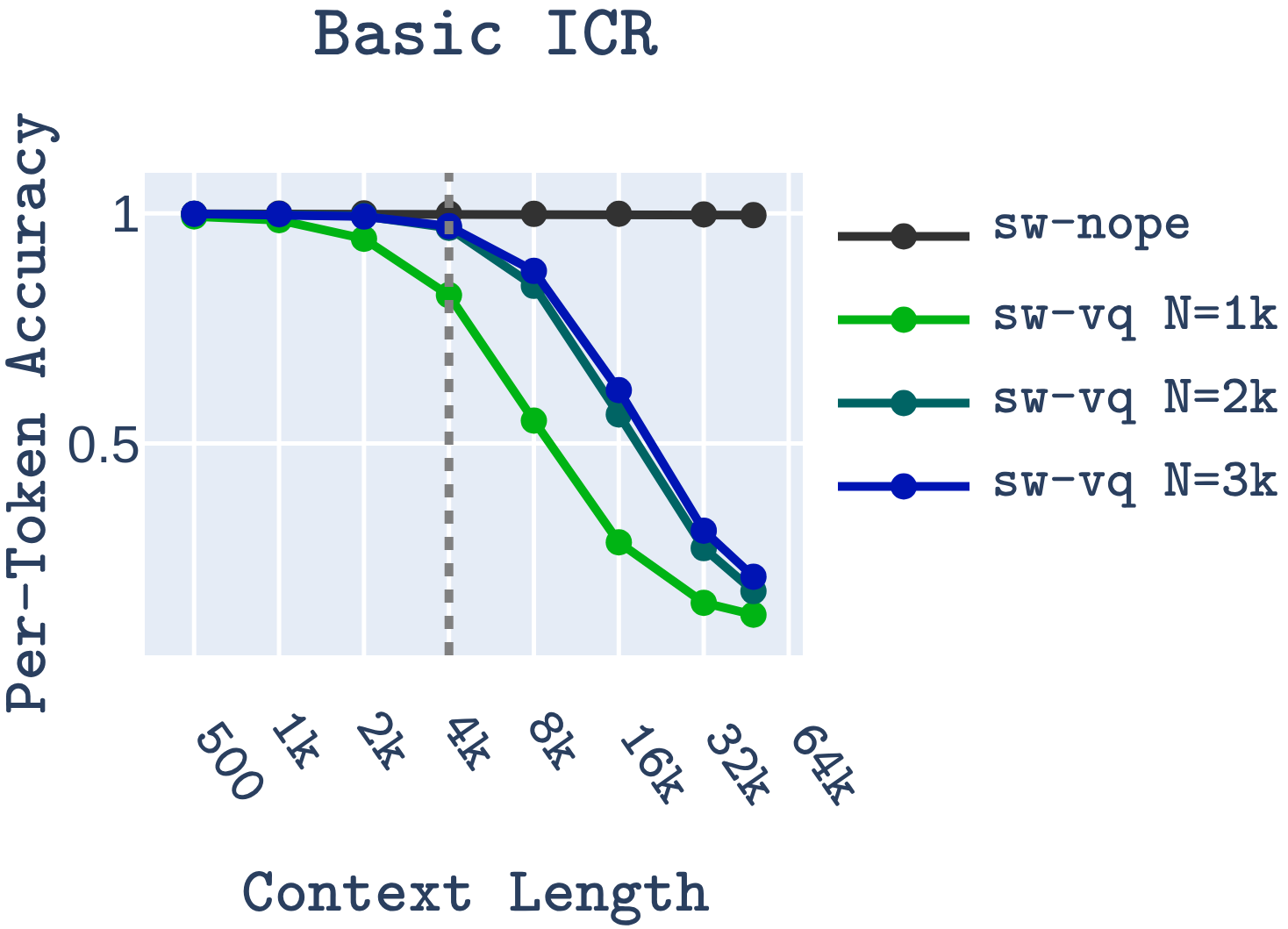}
\caption{\textbf{Preliminary test of in-context recall} in model interleaving sliding window and VQ-attention layers. Differing number of centroids, $\texttt{N}$, are tested.}\label{fig:prelim}
  \end{center}
\end{figure}
Previous works, e.g., \cite{lingleVQ}, tested transformers that utilize VQ-attention layers on short-context language modeling tasks. However, as far as we can find, there has been no direct testing of VQ-attention on tasks requiring long-context abilities. Here we test a model that interleaves sliding window layers with RoPE and VQ-attention layers with NoPE (\texttt{sw-vq}), and we compare to a baseline that interleaves sliding window with RoPE and standard self-attention with NoPE (\texttt{sw-nope}) on a key-value retrieval task (explained further in experiment section). These models are set up so the only difference between the two is that VQ-attention model quantizes keys, while the baseline does not (see equation \ref{eq:vq_quad}). We test the VQ-attention model with dictionaries that have 1k, 2k, and 3k centroids using SOTA dictionary training methods (see appendix \ref{app:dict_prelim} for details). Figure \ref{fig:prelim} shows the results. The baseline model performs the task perfectly up to the maximum training length of 4k and also demonstrates excellent length extrapolation capability, performing nearly perfectly to the maximum test context length of 48k. The performance of the VQ model begins to degrade prior to the maximum train length and shows poor length extrapolation abilities. Importantly, increasing the number of centroids in $D_k$ provides diminishing returns in improvements, meaning this issue cannot be trivially solved by adding more centroids to the dictionaries.

\section{Online Vector Quantized Attention}

The only difference between the self-attention baseline and the VQ-attention model are the quantized keys (see equation \ref{eq:vq_quad}). These performance deficits must therefore be due to the quantization error, i.e., the difference between each $\mathbf{k}_t$ and the nearest neighbor centroid it gets replaced with.\begin{wrapfigure}{r}{0.25\textwidth}
\begin{center}
\includegraphics[width=0.23\textwidth]{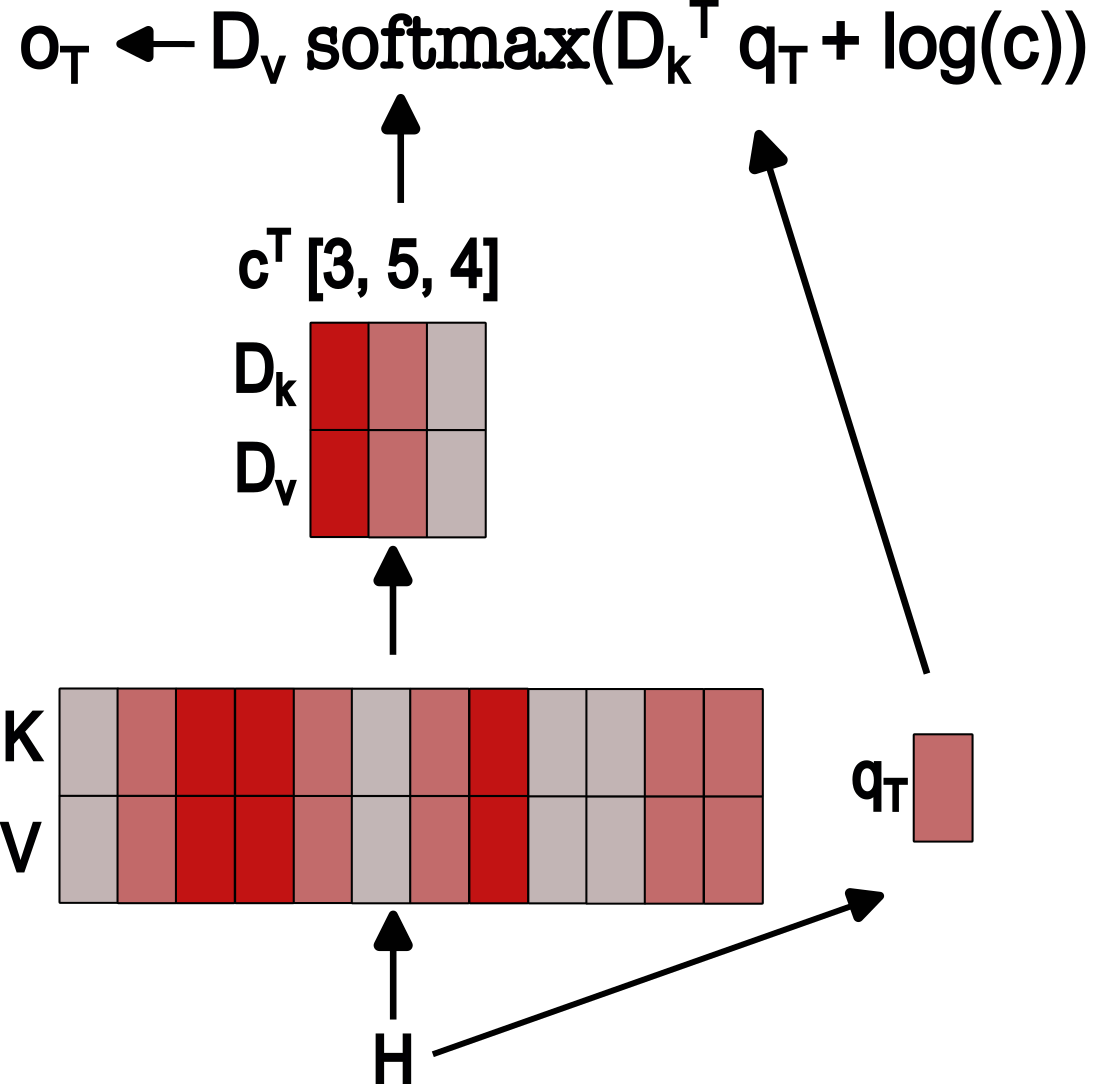}
\caption{Process for generating OVQ-attention output for token at position $T$.}\label{fig:ovq_pred}
  \end{center}
\end{wrapfigure}This quantization error adds a bias to the self-attention operation and the gradients, since we must use a straight-through estimator to propagate gradients through the quantization operation. As we saw, this performance degredation cannot be trivially solved by increasing the size of $D_k$. We, therefore, propose an alternative solution, which is to learn $D_k$ \textit{online} during the forward pass. We call this \textit{online vector quantized attention} (OVQ-attention), since now all parameters of VQ-attention ($D_k$, $D_v$, and $\mathbf{c}$) are learned on-the-fly during the forward pass. By dynamically computing $D_k$ during the forward pass, we hypothesize that $D_k$ should be able to better fit keys to both in and out of training distribution lengths and provide potential benefits by removing the need for a straight through estimator.

\subsection{Gaussian Mixture Regression: A Theoretical Basis for OVQ-attention}
First, we provide a novel theoretical basis on which to develop and further motivate OVQ-attention. Self-attention has previously been shown to be equivalent to Gaussian kernel regression (GKR), under certain assumptions (e.g., \cite{bahdanau2014neural, murphy2022probabilistic}). GKR estimates the conditional probability distribution $P(V|K)$ using samples $(\mathbf{k}_0, \mathbf{v}_0), (\mathbf{k}_1, \mathbf{v}_1), ...,(\mathbf{k}_T, \mathbf{v}_T)$. It's prediction is the expected value of $V$ given a value for input $K$. Assume $K = \mathbf{q}_T$ and queries and keys have the same L2-norm. In this case,
\begin{align}
\mathbb{E}(V|K=\mathbf{q}_T) &= \sum^T_{t=0}\frac{e^{-\beta ||\mathbf{q}_T - \mathbf{k}_t||^2}}{\sum^I_{i=0}e^{-\beta ||\mathbf{q}_T - \mathbf{k}_i||^2}} \mathbf{v}_t\\&= \texttt{softmax}(\beta \mathbf{q}_T K^{\top})V
\end{align}\label{eq:gkr_att}
where $\beta$ is the precision (see appendix \ref{app:GKR_att} for more detailed derivation). This description, combined with equation \ref{eq:vq_quad}, suggests VQ-attention may be interpreted as doing GKR with quantized keys. However, we show VQ-attention has a deeper connection to a different model, which naturally incorporates vector quantization and the count vectors used in the linear form of VQ-attention. In particular, we show VQ-attention, and our OVQ-attention layer in particular, is closely related to a model known as Gaussian \text{mixture} regression (GMR). Like GKR, GMR computes the conditional probability $P(V|K)$ using a dataset consisting of inputs (keys) and outputs (values). Unlike GKR, GMR does not use the raw dataset to compute the conditional distribution. Instead, it first fits a Gaussian mixture model (GMM) to the dataset $[\mathbf{k}_0, \mathbf{v}_0], [\mathbf{k}_1, \mathbf{v}_1], ...,[\mathbf{k}_T, \mathbf{v}_T]$, where the brackets indicate that associated keys and values are concatenated to form a single data point. To generate predictions, the model computes the conditional distribution $P(V | K)$ using some input and the Gaussian mixture. More specifically, assuming that keys and queries have the same L2-norm norm and $\boldsymbol{\Sigma}_n = \mathbf{I}\frac{1}{\beta}$, we show in the appendix \ref{app:GMR_OVQ} that the prediction of a GMR model is 
\begin{align}
\mathbb{E}(V|K=\mathbf{q}_T) &= \sum^N_{n=0}\frac{\mathbf{\pi}_n e^{-\beta ||\mathbf{q}_T - \boldsymbol{\mu}^k_n||^2}}{\sum^J_{j=0} \mathbf{\pi}_j e^{-\beta ||\mathbf{q}_T - \boldsymbol{\mu}^k_j||^2}} \boldsymbol{\mu}^v_n\\&= \texttt{softmax}(\mathbf{q}_T D_k^{\top} + \text{log}(\mathbf{c}_T))D_v,\label{eq:GMR_Inf}
\end{align} 
where $\boldsymbol{\mu}^k_n$ and $\boldsymbol{\mu}^v_n$ are the $n$-th means in $D_k$ and $D_v$, respectively, and $\pi_n$, is the prior probability distribution, which is proportional to the counts, $\mathbf{c}_n$. Although the way GMR and VQ-attention generate predictions is related, the way these models learn is significantly different: VQ-attention learns $D_k$ and $D_v$ separately, one in pretraining and the other online during inference, while GMR learns these simultaneously, treating them as a single set of parameters fitted to the same dataset. This further motivates an alternative form of VQ-attention that learns both $D_k$ and $D_v$ simultaneously: in order to accurately compute $P(V|K)$, the underlying GMM model must be trained in a principled way such that $D_k$ and $D_v$ are fitted to the same dataset of interest, which in the case of sequence mixing layers, is the input sequence of keys and values. We now show how our OVQ-attention accomplishes this learning process.

\begin{figure}[h]
  \begin{center}
\includegraphics[width=0.38\textwidth]{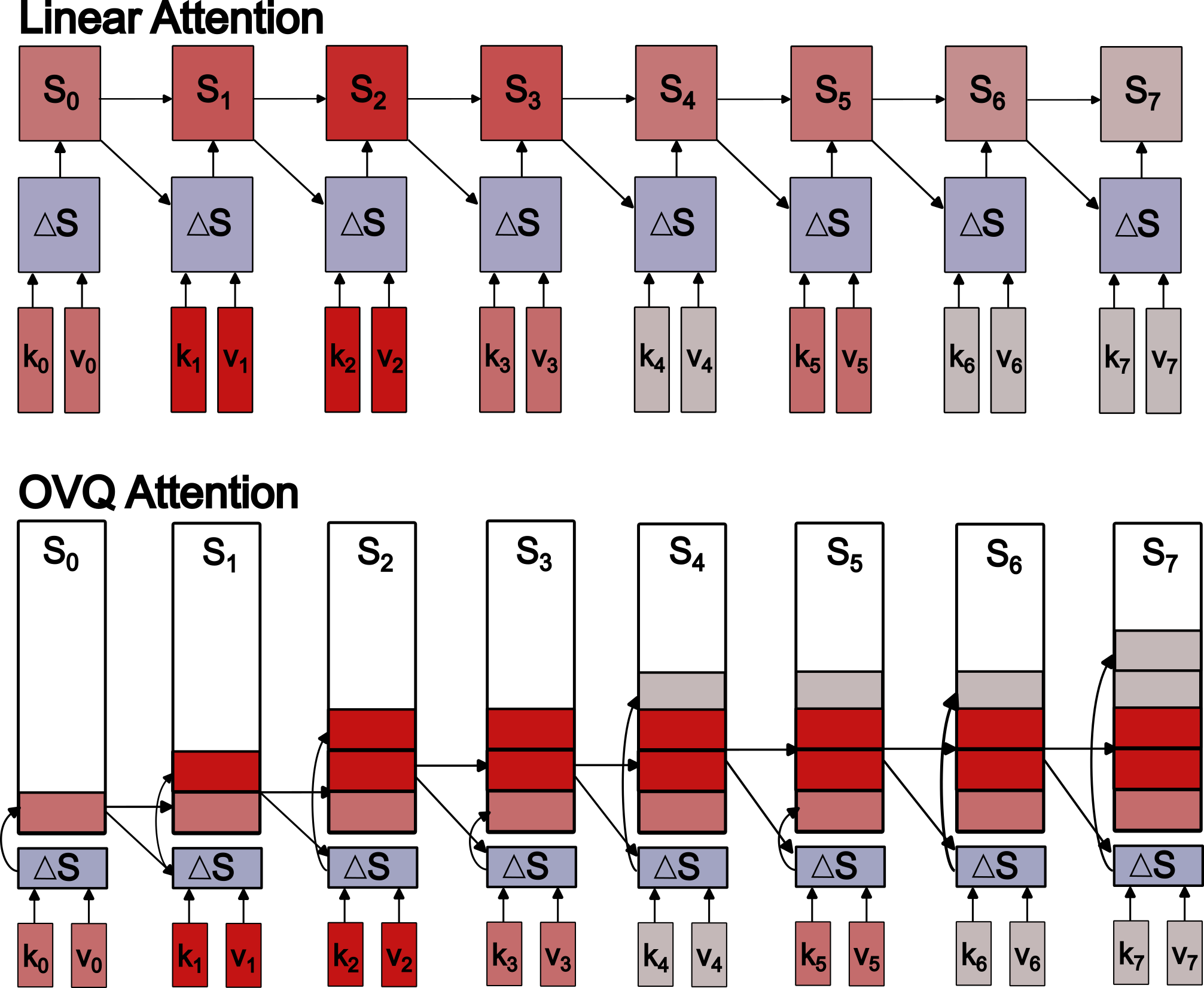}
\caption{State updates in linear and OVQ-attention models.}\label{fig:ovq_learn}
  \end{center}
\end{figure}

\subsection{The Online Learning Formulation}
Following the test time training framework \cite{sun2024learning}, we use an online learning formulation for OVQ-attention. We assume at each point in the sequence, $t$, we have a GMR model that is presented with $\mathbf{q}_t$, $\mathbf{k}_t$, and $\mathbf{v}_t$. The model must generate a prediction, as in equation \eqref{eq:GMR_Inf} above, using $\mathbf{q}_t$ and the current model, and it must update the parameters ($D_k$, $D_v$, and the counts $\mathbf{c}$) using $\mathbf{k}_t$ and $\mathbf{v}_t$ to optimize the loss function for Gaussian mixtures, described below. In practice, we use a chunk-parallel formulation, in which the model is presented with a small chunk of consecutive queries, keys, and values. The model does a mini-batch update and generates predictions so that the update and prediction generation are parallelized across the chunk. As above, we assume $\mathbf{q}_t$ and $\mathbf{k}_t$ are unit norm and $\boldsymbol{\Sigma}_n = \mathbf{I}\frac{1}{\beta}$, where $\beta$ is a parameter fixed during test time training and learned in the outer-loop.
\begin{figure*}[t]
\centering
\includegraphics[width=0.99\textwidth]{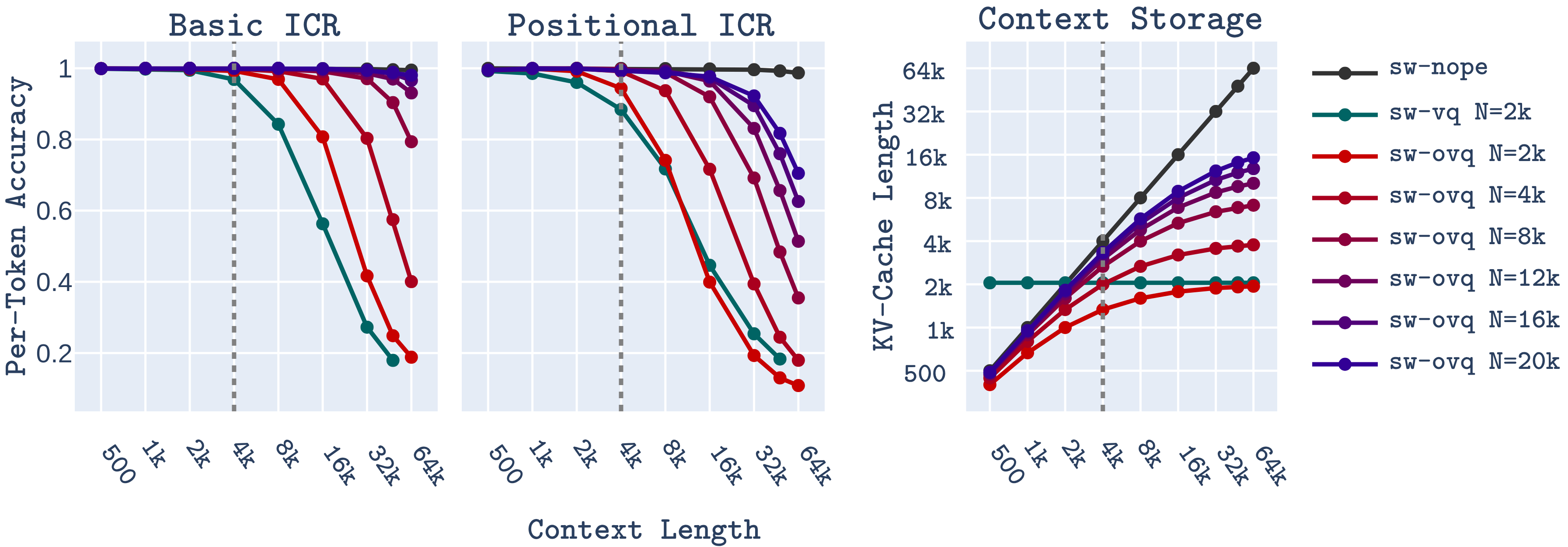}
\caption{\textbf{In-context recall.} Left two plots show per-token-accuracy for our two synthetic recall tasks up to 64k context length. The right plot shows how the memory state, i.e. kv-cache, grows with context length.}\label{fig:ICR}
\end{figure*}

\textbf{Prediction.} Our algorithm uses a chunk-wise parallel implementation, where the model loops through chunks, $Q_c, K_c, V_c$, each of length $L$, but parallelizes operation within each chunk. Every chunk, the model first generates a prediction over $V$ using the expectation $E(V|K=\mathbf{q}_T)$ given an input query via our GMR, shown in equation \eqref{eq:GMR_Inf}. To do this, we first merge the data points in the current chunk with the existing mixture model. Let $D_k^* = [D_k, K_c]$ and $D_v^* = [D_v, V_c]$, where the brackets are concatenations. Let $\mathbf{c}_{c-1}$ be the counts updated to chunk $c-1$, and let $\mathbf{c}^* = [\mathbf{c}_{c-1}, \mathbf{1}]$, where $\mathbf{1} \in \mathbb{R}^{1\times L}$, is a tensor of ones, which act as the counts for the new keys and values concatenated to the dictionaries. 
\begin{equation}
O^{\texttt{ovq-att}} = \texttt{softmax}(\beta Q_c D_k^{*\top} + \text{log}(\mathbf{c}^*) + M) D_v^*,
\end{equation}
where $M$ is a causal mask. Note that unlike the original VQ-attention, the chunks concatenated to the dictionaries are not quantized, but rather are the keys and value in their original form. Further, because both $D_k$ and $D_v$ are weighted sums of the keys and values (see below), gradients can propagate through these matrices without any straight through estimator.

\textbf{Learning.} The loss function minimized during training of GMRs is the negative log likelihood (NLL):
\begin{equation}
    \mathcal{L}(\theta) = -\sum_{t=1}^{T} \ln \left( \sum_{n=1}^{N} \pi_n \mathcal{N}([\mathbf{k}_t, \mathbf{v}_t]  \mid [\boldsymbol{\mu}^k_n, \boldsymbol{\mu}^v_n], \boldsymbol{\Sigma}_n) \right),
\end{equation}
where, in the online setting, $T$ is the current timestep/iteration. That is, this loss sums the NLL of the model over the current and all previously observed data points. In the offline setting, GMMs are typically trained via expectation maximization (EM) \cite{dempster1977maximum}, which has guarantees of converging to minima of the NLL. EM iteratively performs batch updates on the parameters for multiple epochs until convergence. In the online scenario, however, it is assumed there is not enough memory or compute available to store and train on all data points observed up until $T$. Thus, the train set as a whole cannot be retroactively used to perform multiple batch updates using EM. However, in the appendix we show that what can be done in the online setting is to closely approximate 1) the standard initialization process for Gaussian mixtures, and 2) a single batch EM update over these initialized parameters, under our assumption $\boldsymbol{\Sigma}_n = \mathbf{I}\frac{1}{\beta}$. A single batch update may not seem sufficient to achieve a good GMM parameters, but we show that the first EM update in this setting is equivalent to performing a Newton update on the NLL thus providing a good second-order approximation of parameters that minimize the NLL.

A standard method for initializing the means of the $N$ components in a GMM is to set the means equal to a subset of $N$ data points from the train set \cite{bishop2006pattern, pedregosa2011scikit}. This can be done through random sampling, but there are more effective methods. One industry standard, k-means++ \cite{arthur2006km++}, uses a procedure that attempts to maximize the distance between initial Gaussian means. We approximate this process by adding some number, $n_{new}$, components each chunk $c$. To determine the number of new components to add each chunk we use the following growth function for the dictionaries:
\begin{equation}\label{eq:growth}
N_t = t \frac{N}{t + N},
\end{equation}
where $N_t$ is the number of centroids in $D_k$ and $D_v$ at step $t$, $N$ is the maximum number of centroids. This is a plateauing growth function, loosely inspired by the Dirichlet process \cite{escobar1994dirichlet}, that grows quickly early in the sequence and slows later in the sequence. We can see that as $t \rightarrow \infty$, $N_t \rightarrow N$. For every chunk, $c$, of length $L$ we calculate the number of new centroids:
\begin{equation}
n_{new} = N_{Lc} - N_{L(c-1)}.
\end{equation}
The data points within the current chunk assigned as new centroids are the $n_{new}$ that have the smallest dot products with their nearest neighbor centroid. Like k-means++ \cite{arthur2006km++}, this procedure aims to have a large spread between initial centroids. Data points not assigned as new means are merged with old means. We use the following update
\begin{equation}
\Delta [\boldsymbol{\mu}^k_{t^*}, \boldsymbol{\mu}^v_{t^*}] =  \frac{1}{c_{t^*} + c_{t^*,c}} ([\mathbf{k}_t, \mathbf{v}_t]  - [\boldsymbol{\mu}^k_{t^*}, \boldsymbol{\mu}^v_{t^*}]),
\end{equation}\label{eq:online_kmeans}
where $[\boldsymbol{\mu}^k_{t^*}, \boldsymbol{\mu}^v_{t^*}]$ is the nearest neighbor cluster assignment to $[\mathbf{k}_t, \mathbf{v}_t]$ and $c_{t^*,c}$ is the number of datapoint in the current chunk, $c$, assigned to cluster $t^*$. This is an online version of the batch K-means update. We show in the appendix that the first batch EM update performed after initializing parameters is equivalent to batch K-means, under our assumptions about covariance matrix. Importantly, previous results \cite{bottou1994convergence} can be used to show this first EM update is equivalent to a Newton update on the NLL loss (see appendix \ref{app:GMR_OVQ}), which yields a second-order approximation of parameters that minimize the loss.
\begin{figure*}[t]
\centering
\includegraphics[width=0.99\textwidth]{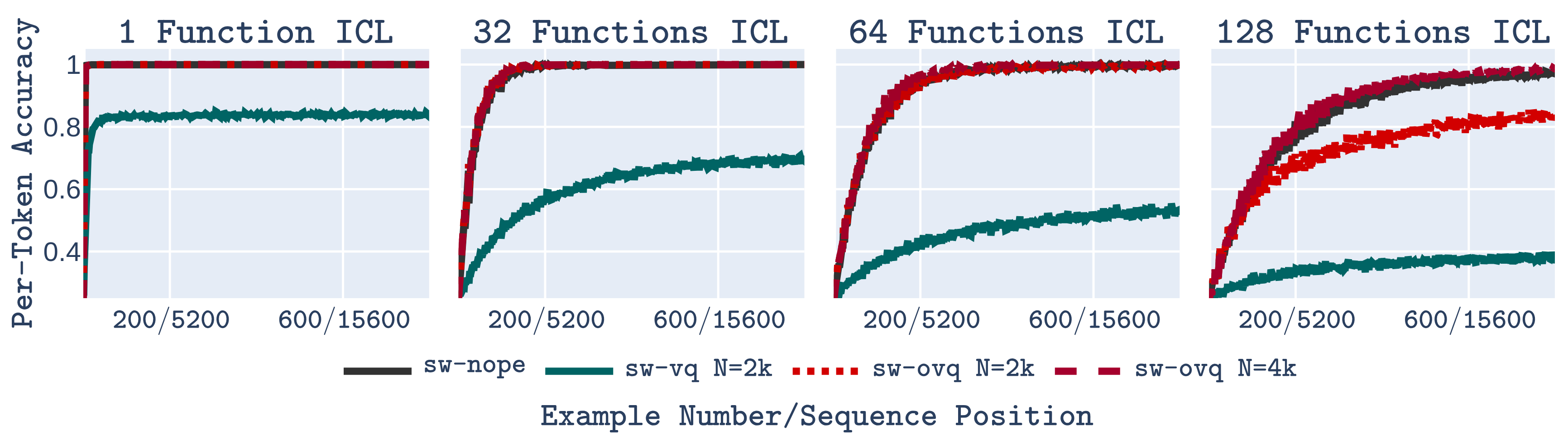}
\caption{\textbf{In context learning}. Models are trained on 2k context with 16 functions. We show the per-token accuracy over the output, $\mathbf{y}_n$, for each example, $n$, in the context, averaged over the test set.}\label{fig:ICL}
\end{figure*}
\subsection{Computational Properties For Training}
OVQ stores $D_k$, $D_v$ and count tensor $\mathbf{c}$. As $T \rightarrow \infty$ these tensors grow in their number of components toward the hard upper bound of $N$, incurring constant $\mathcal{O}(N)$ memory costs. The main compute cost in OVQ-attention comes from the attention operation $Q_c D_k^{*\top}$ and the operation to get dot product values between keys and centroids, $K_c D_k^{\top}$, yielding linear compute cost $\mathcal{O}(2 N L C) = \mathcal{O}(2 N T)$, where $L$ is fixed chunk length and $C$ is the number of chunks. We further describe OVQ attentions compute benefits in appendix D, where we provide an extensive flops analysis and comparison to self attention and delta net.

\subsection{Comparison to Linear Attention Models} A central difference between OVQ-attention and common sequence mixing layers with constant memory, i.e., SSMs \cite{gu2024mamba, dao2024transformers} and linear attention models \cite{qin2022devil}, is that their state update does not grow in size with the memory state size. Let $d_k$ and $d_v$ be key and value dimension, respectively. Linear attention models, for example, store state $S \in \mathbb{R}^{d_k\times d_v}$. In standard chunk-parallel implementations, state updates are produced for each of the $L$ tokens in the current chunk yielding $\Delta S \in \mathbb{R}^{L \times d_k\times d_v}$. The memory footprint of the update tensor limits the chunk size and the state size that can be used in practice. OVQ-attention sparsely updates its state. Each of the $L$ tokens in the current chunk only update a single row of $D_k$ and $D_v$. Thus, only updates for specific rows are computed and stored, yielding $\Delta S \in \mathbb{R}^{L \times 2 \times d}$. Memory updates can therefore be expressed in terms of gather and scatter operations (see appendix \ref{app:ovq_implement}) that only retrieve and operate on a fixed number of columns of the memory state. Importantly, this implies the memory footprint of $\Delta S$ is unrelated to the number of components, $N$, in the dictionary. OVQ-attention can, consequently, increase its state size, $S$, without increasing the memory footprint of $\Delta S$. Increasing $N$ increases compute costs elsewhere but mainly to matrix multiplications (see previous section), which can be highly optimized.

\section{Experiments}
We test OVQ attention on a variety of long-context synthetic and natural language tasks. We focus on hybrid architectures that interleave sliding window attention layers with standard, VQ, and OVQ attention layers with NoPE. We get our best results with these sorts of architectures, both for VQ and standard attention layers. However, we test a version of OVQ attention that incorporates RoPE in appendix C, and find it matches or surpasses standard attention layers with RoPE on several tasks. 
\subsection{Synthetic Long-Context Tasks} Sequence mixing layers with constant memory, like SSMs and linear attention layers, are known to struggle significantly relative to self-attention with tasks that require in-context recall (ICR) and in-context learning (ICL) over long context lengths \cite{akyurekcontextlang, wang2025systematic}. We test our models on several hard synthetic tasks that stress these abilities. All models tested on these tasks have eight layers and model dimension of 768 and 70M parameters.
\begin{figure*}[t]
\centering
\includegraphics[width=0.94\textwidth]{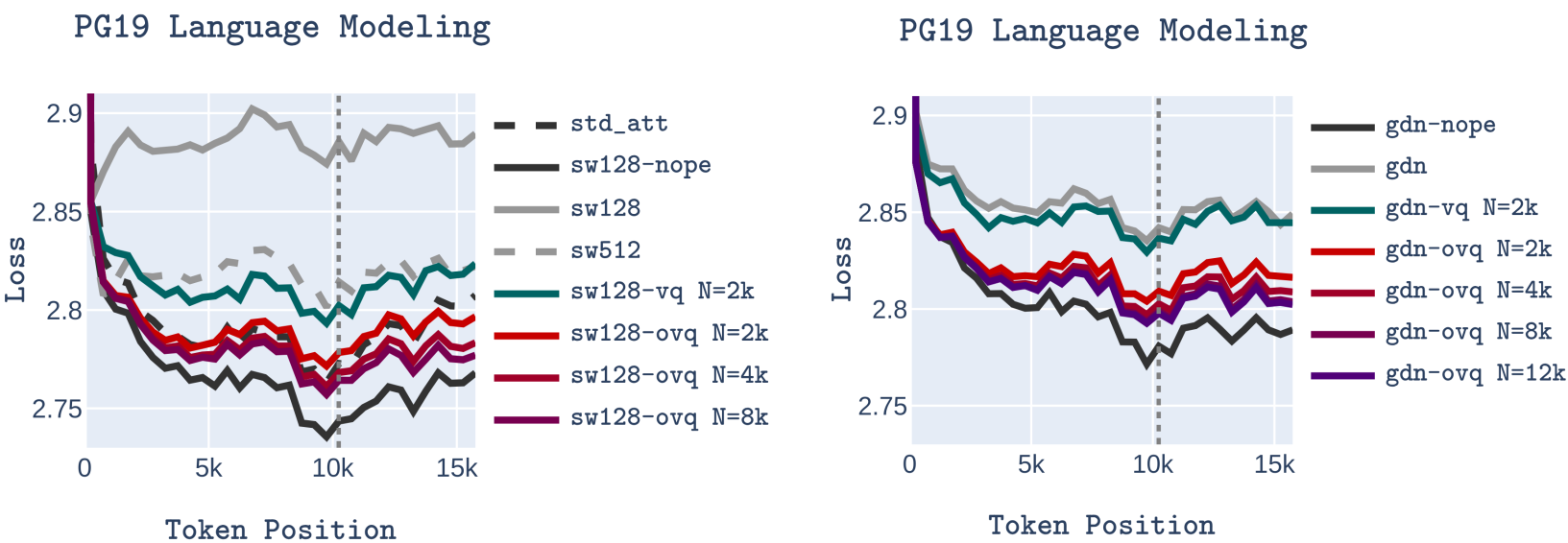}
\caption{\textbf{Long context language modeling on PG19.}}\label{fig:pg19}
\end{figure*}

\textbf{Long In-Context Recall.} We create two synthetic ICR tasks. The first, which we call "basic ICR", fills a context with key-value pairs. Each key and each value are unique and consist of eight tokens. At the end of the context, a random sample of six keys-value pairs from the context are presented, and the model must predict the correct value tokens for these key-value pairs. We measure the per-token accuracy over these value tokens (i.e., the proportion of tokens in each value that is correctly predicted). The second ICR task has the same setup except in the context there are four copies of each key, and each copy is assigned a distinct value. At the end of the context, four copies of one key are presented and the model must output all of its associated values in the order they appeared in the context. We call this "positional ICR", since it requires understanding the global relative position of key-values pairs. Models are trained at context lengths 500, 1k, 2k, and 4k, and tested up to 64k. Results are shown in figure \ref{fig:ICR}. Our strongest baseline, \texttt{sw-nope}, which interleaves sliding window size 128 and full attention NoPE layers, performs both of these tasks almost perfectly up to 64k. The \texttt{sw-vq} layer struggles on these tasks at long train length of 4k, and has quickly decaying performance beyond 4k. We train a single \texttt{sw-ovq} with $N$=2k centroids. However, we test at a variety of larger dictionary sizes to see if the model can adapt at test time. The \texttt{sw-ovq} shows superior performance to \texttt{sw-vq} at similar dictionary sizes (N=2k to 4k), \texttt{sw-ovq}. Further, increasing the maximum dictionary size at test time beyond the train length shows considerable improvements in OVQ-attention layers. On basic ICR, \texttt{sw-ovq} very nearly matches performance up to 64k context length w $N=$16 and $N=$20, which compress the context to 75-80$\%$. Performance is slightly worse on the more difficult positional ICR, but still near perfect performance up to 16k context length for $N \geq 4$k+. Further, we find small architecture tweaks can improve the performance of \texttt{sw-ovq} on positional ICR by 10-20$\%$ at 16k-64k context lengths (see appendix C). Finally, in figure \ref{fig:linear_baselines}, we see equal-parameter linear attention baselines fail at both tasks beyond 2k sequence length.

\textbf{Long In-Context Learning.} Next, we test models on a long-context version of linear regression based ICL tasks \cite{akyurek2022learning}. In this task, the context consists of input-output pairs, $\mathbf{x}, \mathbf{y}$, where the output of each is generated by some linear function, $\text{func}_f(\mathbf{x}) = \mathbf{y} = b + a P \mathbf{x}$. In our tests, $a$ and $b$ are integers such that $0 < a < 5$,  $0 < b < 5$ and $P$ is a permutation matrix. Using a permutation matrix and small $a$ and $b$ ensures the function output is a vector of integers within the model's vocab range. Each input-output pair has 12 tokens. To make this task require ICL over long-contexts, input-output pairs from multiple functions are shown in context. Each example is marked with a special token that identifies which function type generated the output. This means the more function types there are in context the greater the average distance between examples for each function, $\text{func}_f$. For example, we find it takes transformers about five examples, minimum, to learn a function. In our hardest test, with 128 functions, the average distance between input-output examples for some function $\text{func}_f$ is 3000+ tokens, so information must be integrated, on average, over 15000+ tokens to learn each function. Models are trained on context lengths of 2k with 16 functions and tested long contexts with 1, 32, 64, and 128 functions. Figure \ref{fig:ICL} displays the results. The baseline \texttt{sw-nope} model successfully learns all functions in all scenarios, even in the hardest case 128 function scenario. The \texttt{sw-ovq} model impressively matches the performance of \texttt{sw-nope}, even when using a small dictionary $N=$4k. The \texttt{sw-vq} model struggles to successfully learn even a single function in this task. Equal-parameter linear attention baselines also struggle to perform both ICL and basic ICR beyond 2k contexts (see figure \ref{fig:linear_baselines}).

\subsection{Long Context Language Modeling} We also test OVQ models on long context language modeling using the PG19 dataset. We filter the dataset to remove sequences less than 16k tokens. Sliding window and gated delta net (\texttt{gdn}) \cite{yang2024gated} interleaved models are tested at size 230M parameters each. Models are trained at sequence length 10k and tested at 16k sequence length. OVQ-attention layers are trained with $N=$2k. Results are shown in figure \ref{fig:pg19}. Adding OVQ layers to pure sliding window and \texttt{gdn} models drastically improves long context language modeling performance, even for small $N=2$k. Performance of \texttt{gdn-ovq} and \texttt{sw-ovq} does not quite match the nope baseline at long lengths, differing at 10k by about .02 cross-entropy. We found our \texttt{sw-nope} interleave baseline to be significantly stronger than the full attention w/ RoPE baseline, \texttt{std-att} on long context language modeling. The \texttt{sw-ovq} model matches, even slightly surpasses, \texttt{std-att}.
\begin{figure}[h]
\includegraphics[width=0.48\textwidth]{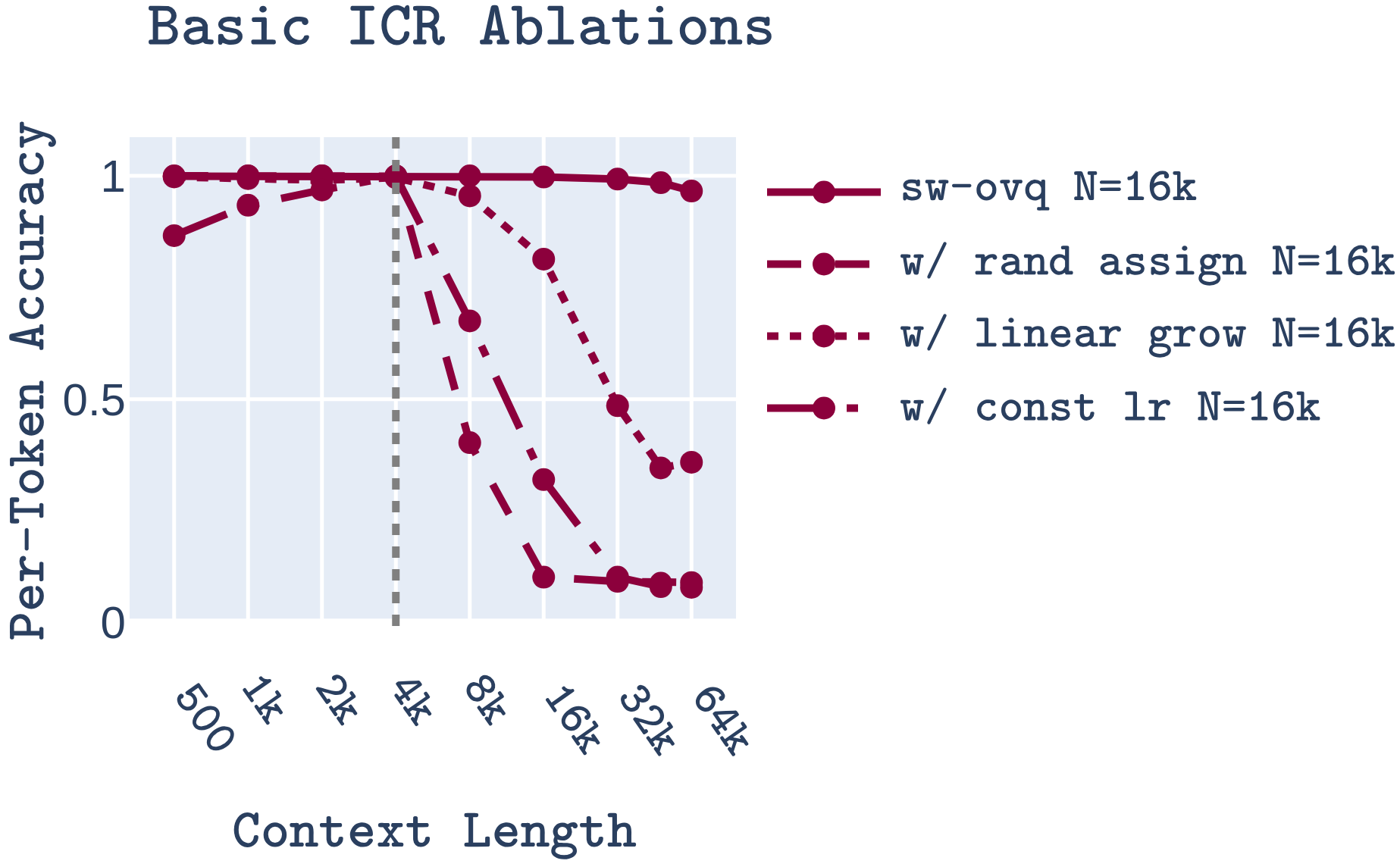}\caption{\textbf{Ablations on Basic ICR.}}\label{fig:ablate1}
\end{figure}
\begin{figure*}[t]
\centering
\includegraphics[width=0.99\textwidth]{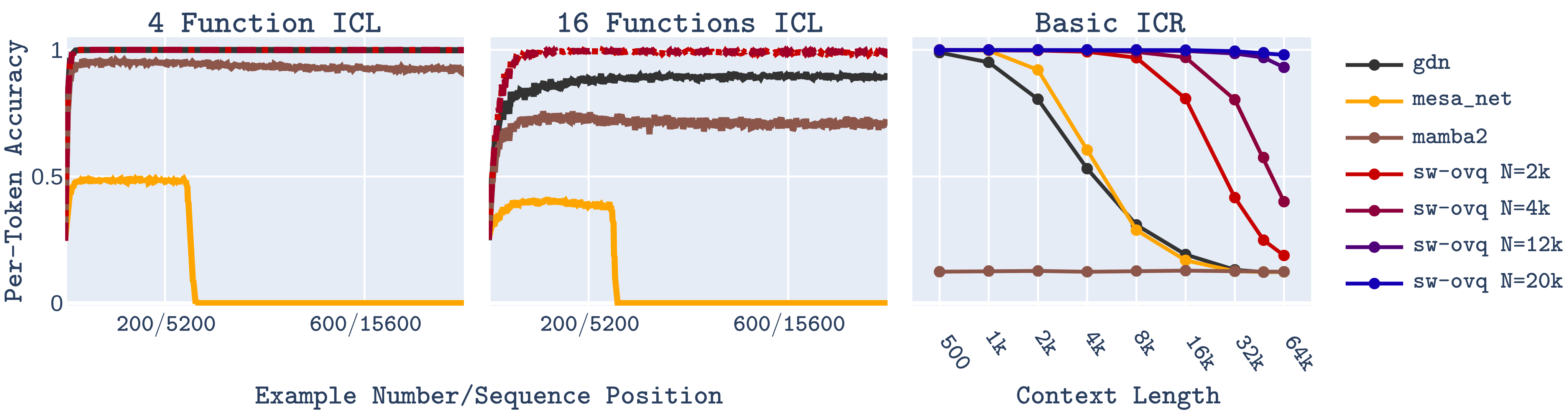}
\caption{\textbf{Comparison to Linear Attention and SSM Baselines.} Left two plots show per-token-accuracy for ICL tasks. Models train on 2k context with four functions, tested on four and 16 functions. The right plot shows performance for basic ICR.}\label{fig:linear_baselines}
\end{figure*}
\begin{table*}[t]
    \centering
    \begin{tabular}{lccccccc} 
        \toprule
        \textbf{Model} & \textbf{Params} & \textbf{PIQA} & \textbf{Hella.} & \textbf{Wino.} & \textbf{ARC-e} & \textbf{ARC-c} & \textbf{Avg.} \\ 
        \midrule
        \texttt{std att} & 480M & $66.4^{\pm0.1}$ & $40.5^{\pm0.1}$ & $52.3^{\pm0.7}$ & $52.8^{\pm0.4}$ & $29.1^{\pm0.5}$ & $48.21$ \\
        \texttt{sw-nope} & 480M & $66.7^{\pm0.4}$ & $40.9^{\pm0.1}$ & $52.6^{\pm0.6}$ & $53.1^{\pm0.1}$ & $28.4^{\pm0.4}$ & $48.35$ \\
        \texttt{sw-ovq} & 480M & $66.6^{\pm0.2}$ & $41.1^{\pm0.2}$ & $52.4^{\pm0.2}$ & $52.7^{\pm0.1}$ & $28.7^{\pm0.4}$ & $48.30$ \\
        \bottomrule
    \end{tabular}
    \caption{Short context benchmarks.}
    \label{tab:my_data}
\end{table*}

\subsection{Short Context Benchmarks}
OVQ-attention does not compress keys and values significantly over short contexts, so we should expect similar performance on common short context benchmarks to standard attention. To confirm this, we compare a 480M parameter \texttt{sw-nope}, \texttt{std att}, and \texttt{sw-ovq} model on standard short context benchmarks. Models are trained for 50B tokens of \texttt{fineweb-edu} \cite{penedo2024fineweb} at context length 1536. Scores shown in table \ref{tab:my_data} are the averages and standard deviation over the last three checkpoints of the training run for each model and test. We see nearly all scores are within one standard deviation of each other, and the average score for \texttt{sw-ovq} is well within the average standard deviation levels of the other two baselines. 

\subsection{Ablations}
We test several ablations on our OVQ layer. First, instead of setting new centroids using our spread maximizing scheme, we set new centroids to a random sample of key-values from the current chunk. Second, instead of using our plateauing dictionary growth method, we test adding new centroids linearly (i.e., dividing $n_{new}$ evenly across each chunk). Finally, instead of using the adaptive learning rates for updating dictionary centroids (equation \ref{eq:online_kmeans}), we use a constant learning rate. Using a constant learning rate makes the dictionary update equivalent to a kind of gradient descent, or first order approximation of the minimizing parameters rather than a second-order one. We test on basic ICR (figure \ref{fig:ablate1}) and PG19 language modeling (appendix C). Each ablation worsens performance in at least one of these tasks.

\section{Related Works}

\textbf{Previous work on VQ-attention.} VQ-attention was first proposed by \citet{lingleVQ}. In this initial paper, a model with VQ-attention was tested on relatively short sequence language modeling. We could only find two subsequent works, \cite{li2025vql, huang2025vqt}, that try to build and improve upon original VQ-attention. Both apply product quantization and/or residual quantization methods to try to reduce the quantization error. However, these modifications are not compatible with linear compute and constant memory. Although they can reduce memory and compute, relative to standard attention, they yield the same complexity. Our aim, however, is to improve performance of VQ-attention, while maintaining its compute and memory complexity.

\textbf{VQ for KV-cache Compression and Sparse Attention.} Multiple works use vector quantization (e.g., \cite{kumar2024residual, li2025commvq, li2025antkv}) as a method for KV-cache compression applied after training. While these methods lead to a reduction KV-cache memory cost, they do not reduce the complexity of memory or compute compared to standard self-attention. VQ has also been used to make sparse attention operations more efficient (e.g., \cite{zhang2025pqcache, hang2025enhancing, liu2025clusterkv, zhang2025clusterattn}). These methods often achieve linear complexity, however they do not provide a method for storing keys and value using constant memory.

\textbf{Fast Product Key Memory.} Perhaps most similar to our OVQ-attention is the recent fast product key memory (FPKM) layer \cite{zhao2026fast}. This algorithm was published during the writing of this manuscript and its code has not been released, so comparisons could not be made. FPKM uses a sparse top-k update on a constant sized memory. Unlike OVQ-attention, FPKM does not use multiple heads and does not have principled adaptive learning rates or dictionary growth.

\section{Discussion and Limitations}

\textbf{Limitations.} Our OVQ layer was tested at relatively small scale ($<500$M), so more testing is needed to understand how it performs at larger scales. A hardware-efficient implementation of OVQ-attention has yet to be developed, so it remains to be seen how a highly optimized implementation of OVQ attention compares to those of linear and self attention in terms of latency and throughput.

\textbf{Future Directions and Continual Learning.} Beyond scaling and improving the efficiency of OVQ, we believe there is promise to investigating the use of OVQ-attention layers for continual learning. OVQ-attention fits within the test-time training framework \cite{sun2024learning}, which formalizes linear attention layers as storing a model that optimizes an online loss function. The best performing linear attention models implement linear regression layers or MLPs that use types of dense, gradient updates such as  \cite{sun2024learning, yang2024gated, behrouz2024titans}. In language modeling, it is typical that individual input sequences are non-I.I.D, implying the online learning problem these models must solve to achieve ICL is thus also a continual, non-I.I.D. learning problem. It is well-established that MLP-like models trained with gradient descent suffer from catastrophic forgetting in non-I.I.D. learning scenarios \cite{mccloskey1989catastrophic, french1999catastrophic}. This fact may partially explain why linear attention and similar SSM models often struggle on long-context tasks. OVQ-attention, on the other hand, implements an associative memory mechanism more closely related to online Gaussian mixture models and online clustering. The sparse updates used in online clustering allow for fast learning, while being resilient to catastrophic forgetting. For example, \citet{alonso2024sparse} showed, in a variety of non-I.I.D., online-continual learning scenarios, that associative memory models trained via online clustering-like algorithms are able to rapidly adapt to data distribution shifts without catastrophic forgetting. In figure \ref{fig:ovq_learn}, we visualize how the state updates performed by linear attention models may lead to in-context catastrophic forgetting, while the sparse OVQ-attention updates prevent forgetting. OVQ-attention, in this way, may provide a useful basis for rapidly incorporating new knowledge and skills into LLMs without catastrophic forgetting. Future work could look into explicitly testing the continual learning capabilities of such layers.

\textbf{Conclusions.} OVQ-attention, like linear attention and SSMs, has constant memory and linear compute complexity. Unlike linear attention and SSMs, OVQ-attention uses sparse state updates that are small and independent of state size (along a certain dimension). State size can, therefore, be increased without increasing the memory footprint of the state update, leaving room to use very large state sizes that drastically increase memory capacity. We provided empirical evidence that OVQ-attention has drastically improved ICR, ICL, and language modeling at long context lengths relative to linear attention and SSMs. This approach, therefore, marks an important research direction for efficient alternatives to self-attention.

\section*{Acknowledgements}
Thanks to Vasu Shyam, Kamesh Krishnamurthy, Anna Golubeva, Leon Lufkin, Manos Theodosis and the rest of the Zyphra team for useful discussions and comments on earlier drafts of the paper. 

\bibliographystyle{icml2026}
\bibliography{references}

\newpage
\onecolumn

\section{Appendix A: Theoretical Results}

\subsection{Details on Relation between Gaussian Kernel Regression and Self-Attention}\label{app:GKR_att}
Assume $\mathbf{q}_t$ and $\mathbf{k}_t$ are unit norm. In this case,
\begin{align}
-\frac{1}{2}||\mathbf{q}_T - \mathbf{k}_t||^2 &= -\frac{1}{2}(||\mathbf{q}_T||^2 + ||\mathbf{k}_t||^2  - 2  \mathbf{q}_T \mathbf{k}_t^{\top})\\
&= -\frac{1}{2}(1  + 1  - 2  \mathbf{q}_T \mathbf{k}_t^{\top})\\
&= -1 + \mathbf{q}_T \mathbf{k}_t^{\top}.
\end{align}
If we take the exponent of this result we get
\begin{align}
e^{-1 + \mathbf{q}_T \mathbf{k}_t^{\top}} = P  e^{ \mathbf{q}_T \mathbf{k}_t^{\top}}
\end{align}
where $P$ is a constant. This result can be used to derive self-attention from Guassian kernel regression \cite{bahdanau2014neural, murphy2022probabilistic}, in the case where $\mathbf{q}_T$ and $\mathbf{k}_t$ have the same norm.
\begin{align}
\mathbb{E}(V|K=\mathbf{q}_T) &= \sum^T_{t=0}\frac{e^{-\frac{\beta}{2} ||\mathbf{q}_T - \mathbf{k}_t||^2}}{\sum^T_{i=0}e^{-\frac{\beta}{2} ||\mathbf{q}_T - \mathbf{k}_i||^2}} \mathbf{v}_t\\ &= \sum^T_{t=0}\frac{e^{\beta  \mathbf{q}_T \mathbf{k}_t^{\top} }}{\sum^T_{i=0}e^{\beta \mathbf{q}_T \mathbf{k}_i^{\top}}} \mathbf{v}_t \\ &= \texttt{softmax}(\beta \mathbf{q}_T K^{\top})V
\end{align}

\subsection{Introduction to Gaussian Mixture Regression}\label{app:GMR_overview}
Here we provide a brief description of GMR based on the description in \cite{stulp2015many}. GMR involves fitting a Gaussian mixture model (GMM) to a dataset, where each data point consists of an input, $\mathbf{k}_t$, and a target output, $\mathbf{v}_t$ combined as $[\mathbf{k}_0, \mathbf{v}_0], [\mathbf{k}_1, \mathbf{v}_1], ...,[\mathbf{k}_T, \mathbf{v}_T]$, where the brackets indicate concatenation. GMR fits a Gaussian mixture model to the joint distribution over these data points:
\begin{equation}\label{eq:joint_gmr_app}
    p(K, V) = \sum_{n=1}^{N} \pi_n \mathcal{N}([K, V]  \mid [\boldsymbol{\mu}^k_n, \boldsymbol{\mu}^v_n], \boldsymbol{\Sigma}_n), \text{   with} \sum_{n=1}^N \pi_n = 1,
\end{equation}
where the prior probability of component $n$ is $\pi_n$, and $[\boldsymbol{\mu}^k_n, \boldsymbol{\mu}^v_n]$ is the concatenation of the mean for the input and output portion of component $n$. The covariance is defined as
\begin{equation}
    \boldsymbol{\Sigma}_n = \begin{pmatrix}
\boldsymbol{\Sigma}^{k}_n & \boldsymbol{\Sigma}^{kv}_n \\
\boldsymbol{\Sigma}^{vk}_n & \boldsymbol{\Sigma}^{v}_n.
\end{pmatrix}
\end{equation}
Generating predictions of output given some new input requires computing the conditional expectation over $V$ given an input using the conditional distribution $P(V | K=\mathbf{q}_T)$ where $\mathbf{q}_T$ is a query input at time $T$. In this case the conditional mean is defined
\begin{equation}
\boldsymbol{\mu}^{v|\mathbf{q}_t}_n = \boldsymbol{\mu}^{v}_n + \boldsymbol{\Sigma}^{vk}_n \boldsymbol{\Sigma}^{k, -1}_n (\mathbf{q}_T - \boldsymbol{\mu}^k_n). \label{eq:conditional_mean}
\end{equation}
The expectation is then the weighted average over each conditional mean, weighted by their conditional probability, computed using the inputs $\mathbf{q}_T$ and input (key) portion of the Gaussian components: 
\begin{equation}\label{eq:joint_gmr_app}
    \mathbb{E}(V|K=\mathbf{q}_T) = \frac{\pi_n \mathcal{N}(\mathbf{q}_T \mid \boldsymbol{\mu}^k_n, \boldsymbol{\Sigma}^{k}_n)}{\sum_{i=1}^{N} \pi_i \mathcal{N}(\mathbf{q}_T  \mid \boldsymbol{\mu}^k_i, \boldsymbol{\Sigma}^{k}_i)} \boldsymbol{\mu}^{v|\mathbf{q}_t}_n
\end{equation}
GMR models are typically trained using expectation maximization (EM) \cite{dempster1977maximum}. GMRs use EM to fit a Gaussian mixture model to maximize the likelihood of the data, P(K,V), or equivalently minimize the negative log likelihood (NLL) of the data, where input and outputs are treated as a single data point: 
\begin{equation}
    \mathcal{L}(\theta) = -\sum_{i=1}^{T} \ln \left( \sum_{n=1}^{N} \pi_n \mathcal{N}([\mathbf{k}_t, \mathbf{v}_t]  \mid [\boldsymbol{\mu}^k_n, \boldsymbol{\mu}^v_n], \boldsymbol{\Sigma}_n) \right),
\end{equation}
where $n$ is the component and $\pi_n$ is its prior probability. EM is the standard for minimizing this loss function in the batch training scenario. EM minimizes the NLL over the batch using multiple training iterations. Each iteration, the EM algorithm performs two-steps to achieve its parameter update. 

1. E-Step (Expectation). Evaluate the responsibilities $z_{t,n}$ using the current parameter values. This represents the posterior probability that data point $[\mathbf{k}_t, \mathbf{v}_t]$ belongs to component $n$:
\begin{equation}
    z_{t,n} = \frac{\pi_n \mathcal{N}([\mathbf{k}_t, \mathbf{v}_t]  \mid [\boldsymbol{\mu}^k_n, \boldsymbol{\mu}^v_n], \boldsymbol{\Sigma}_n)}{\sum_{j=1}^{n} \pi_n \mathcal{N}([\mathbf{k}_t, \mathbf{v}_t]  \mid [\boldsymbol{\mu}^k_n, \boldsymbol{\mu}^v_n], \boldsymbol{\Sigma}_n)}.
\end{equation}\label{eq:e_step_app1}
Under our assumptions about the covariance matrix
\begin{equation}
    \mathcal{N}([\mathbf{k}_t, \mathbf{v}_t]  \mid [\boldsymbol{\mu}^k_n, \boldsymbol{\mu}^v_n], \boldsymbol{\Sigma}_n) \propto e^{- \beta ||[\mathbf{k}_t, \mathbf{v}_t]  -[\boldsymbol{\mu}^k_n, \boldsymbol{\mu}^v_n] ||^2}.
\end{equation}\label{eq:e_step_app2}

2. M-Step (Maximization). Re-estimate the parameters using the responsibilities calculated in the E-step.
First, calculate the total weight assigned to component $k$:
\begin{equation}
    \gamma_n = \sum_{t=1}^{T} z_{t,n}
\end{equation}
Then, update the means and mixing coefficients:
\begin{align}
    [\boldsymbol{\mu}^k_n, \boldsymbol{\mu}^v_n]^{\text{new}} &= \frac{1}{\gamma_n} \sum_{t=0}^{T} z_{t,n} [\mathbf{k}_t, \mathbf{v}_t]\\
    \pi_n^{\text{new}} &= \frac{\gamma_n}{\gamma},
\end{align}
where $\gamma = \sum_n \gamma_n$.

\subsection{Linking GMR and OVQ-attention}\label{app:GMR_OVQ} 
\textbf{Assumptions. }
There are two main assumptions on which the following derivations are based:
\begin{itemize}
    \item First, we use a simplified approximation of the covariance, $\boldsymbol{\Sigma}_n = \mathbf{I}\frac{1}{\beta}$ for all $n$, where $\beta$ is a scalar and $\mathbf{I}$ is the identity matrix. Simplifying covariance matrices, in this way, is a common approximating assumption \cite{bishop2006pattern}. An implication of this assumption most important for our results below is that $\boldsymbol{\Sigma}^{vk}_n \boldsymbol{\Sigma}^{k, -1}_n (\mathbf{q}_T - \boldsymbol{\mu}^k_n) = 0$,  since $\boldsymbol{\Sigma}^{vk}_n$ is a zero matrix. Applying this result to equation \ref{eq:conditional_mean}, we get $\boldsymbol{\mu}^{v|\mathbf{q}_t}_n = \boldsymbol{\mu}^{v}_n$.

    \item Finally, we assume $\mathbf{q}_T$ and $\boldsymbol{\mu_n^k}$ are unit norm.
\end{itemize}

\textbf{Setup.} The standard method for training GMR models is EM \cite{dempster1977maximum}, which, as noted above, assumes a batch setting, where the model performs multiple epochs of batch updates. However, we are interested in the online setting, where a single data point, or very small mini-batch of data points, streams in each training iteration, the model only does one pass through the dataset, and there is not enough memory to store all or most the training data to do multiple iterations of batch EM updates. Common online GMM learning algorithms often resemble online k-means (e.g., \cite{hall2005method, declercq2008online}). K-means is identical to EM in the case of uniform prior distribution, and a hard, one hot posterior distribution, which naturally results from using infinite precision, $\beta = \infty$) \cite{bishop2006pattern}. Below, we show how learning in our OVQ model approximates the initialization and one batch EM applied to a GMR model. We then show how this leads to the equivalence between GMR and OVQ-attention predictions. More specifically,
\begin{enumerate}
    \item First, we show how our algorithm implements standard GMM initialization techniques.
    \item Next, we show that, under this initialization scheme and our assumption, $\boldsymbol{\Sigma}_n = \mathbf{I}\frac{1}{\beta}$, the first batch EM step on a GMM is equivalent to a batch k-means update, and we explain how our update rule is the standard online approximation of batch k-means.
    \item Next, we show that the first EM update, after our initialization scheme, is equivalent to a Newton update on the NLL, thus providing a second-order approximation of the minimizing parameters.
    \item Finally, we show that generating predictions using GMR after initialization and one EM update is equivalent to OVQ-attention.
\end{enumerate}

\textbf{1. Initialization.} First, our online algorithm instantiates a standard initialization process for GMMs. EM describes how to update parameters of a GMM, but it does not specify how to initialize parameters of GMMs. The common method for initialization in the batch setting sets the means of the $N$ Gaussian components of a GMM equal to a subset of $N$ data points \cite{bishop2006pattern, pedregosa2011scikit}. This practice produces much better results than random initialization. These $N$ data points may be drawn randomly, or a more strategic algorithm may be used. One standard approach is k-means++ \cite{arthur2006km++}, which chooses a subset of data points to initialize the means with the goal of maximizing the squared distance between the initial means. Our online algorithm, as explained above, initializes means with data points. Our theoretical results only depend on means being initialized with data points (see next step). However, our algorithms also tries to choose these means in a way increases the spread between initial means, similar to k-means++ (see pseudo-code below).

\textbf{2. OVQ-attention Approximates a Batch EM Update.} After initializing means with a sample of $N$ data points, a step of EM is performed using the remaining $T - N$ data points. To perform this EM step the values of the prior distribution, $\pi$, and precision, $\beta$, need to be set. Setting the prior to a uniform distribution is justified at this point, since immediately after initialization, each existing mean has exactly one data point assigned to it, and thus $c_n=\gamma_n=1$ for all $n$ components, yielding a uniform distribution. Additionally, setting the precision $\beta = \infty$ is justified, since, immediately after initialization, the mean of each Gaussian component equals the one data point assigned to it, so there is no empirical variance between mean and its assigned data point. Prior work has shown that a step of K-means is equivalent to a step of EM on a GMM, when the GMM's prior is uniform and variance $\beta = \infty$ \cite{bishop2006pattern} (this can easily be inferred as well from EM description above). Thus, the first step of batch EM after initialization is equivalent to a batch K-means update. Our online update shown in equation \ref{eq:online_kmeans} is the online version of the batch K-means update, in the sense the two are equivalent, under the same cluster assignments \cite{bottou1994convergence}. In practice, cluster assignments will vary between batch and online versions, since in the online version input $\mathbf{k}_t$ will be assigned to a version of the dictionary that is altered by the previous data points, whereas the batch version assigned all data points to the same dictionary after initialization. However, these deviations in cluster assignments are not observed empirically to negatively effect performance negatively, e.g., see \cite{bottou1994convergence}.

\textbf{3. K-means and Newton's Method.} It has previously been establish by Bottou et al. \cite{bottou1994convergence} that a batched K-means update is equivalent to performing a step of Newton's method on the K-mean's loss function, which in our case, would be
\begin{equation}
    \mathcal{L}(\theta) = \sum_{i=1}^{T} ||[\mathbf{k}_t, \mathbf{v}_t] -[\boldsymbol{\mu}^k_{t^*}, \boldsymbol{\mu}^v_{t^*}]||^2,
\end{equation}
where $[\boldsymbol{\mu}^k_{t^*}, \boldsymbol{\mu}^v_{t^*}]$ is the nearest neighbor mean for data point $[\mathbf{k}_t, \mathbf{v}_t]$. Newton's method is equivalent to computing a second-order Taylor approximation of the parameters that minimizes this loss \cite{pedregal2004introduction}. As mentioned above, an EM update on a Gaussian mixture is equivalent to a K-means update in the case of a uniform prior and $\beta = \infty$ \cite{bishop2006pattern}. Additionally, the K-means loss, in this case, is equivalent to the EM loss, i.e., the NLL: 
\begin{align}
    \mathcal{L}(\theta) &= -\sum_{i=1}^{T} \ln \left( \sum_{n=1}^{N} \pi_n \mathcal{N}([\mathbf{k}_t, \mathbf{v}_t]  \mid [\boldsymbol{\mu}^k_n, \boldsymbol{\mu}^v_n], \boldsymbol{\Sigma}_n) \right) \\ 
    &= -\sum_{i=1}^{T} \ln \left( \sum_{n=1}^{N} \mathcal{N}([\mathbf{k}_t, \mathbf{v}_t]  \mid [\boldsymbol{\mu}^k_n, \boldsymbol{\mu}^v_n], \boldsymbol{\Sigma}_n)\right)\\
    &= -\sum_{i=1}^{T} \ln \left( \sum_{n=1}^{N} e^{- \beta ||[\mathbf{k}_t, \mathbf{v}_t]  -[\boldsymbol{\mu}^k_n, \boldsymbol{\mu}^v_n]||^2}\right)\\
    &= -\sum_{i=1}^{T} \ln \left( e^{- \beta ||[\mathbf{k}_t, \mathbf{v}_t]  -[\boldsymbol{\mu}^k_{t^*}, \boldsymbol{\mu}^v_{t^*}]||^2}\right)\\
    &\propto \sum_{i=1}^{T} ||[\mathbf{k}_t, \mathbf{v}_t] -[\boldsymbol{\mu}^k_{t^*}, \boldsymbol{\mu}^v_{t^*}]||^2.
\end{align}
The first step is a result of the fact that the prior is uniform. The second step applies our covariance assumption. The third step results from setting $\beta = \infty$, yielding hard assignments. Thus, a step of EM in the case of a uniform prior and $\beta = \infty$ is equivalent to performing a Newton update on the NLL. 

\textbf{4. OVQ-attention and GMR Prediction.}
Under the assumptions, we can see that the procedure for computing $\mathbb{E}(V|K=\mathbf{q}_T)$ for a GMR is equivalent to OVQ-attention.  
\begin{align}\label{eq:gmr_to_vq}
    \mathbb{E}(V|K=\mathbf{q}_T) &= \frac{\pi_n \mathcal{N}(\mathbf{q}_T \mid \boldsymbol{\mu}^k_n, \boldsymbol{\Sigma}^{k}_n)}{\sum_{i=1}^{N} \pi_i \mathcal{N}(\mathbf{q}_T  \mid \boldsymbol{\mu}^k_i, \boldsymbol{\Sigma}^{k}_i)} \boldsymbol{\mu}^{v|\mathbf{q}_t}_n\\
    &= \sum_{n=0}^{N}\frac{\pi_n  e^{- (\mathbf{q}_T - \boldsymbol{\mu}^k_n)^{\top} \boldsymbol{\Sigma}^{k, -1}_n  (\mathbf{q}_T - \boldsymbol{\mu}^k_n)}} {\sum_{j=0}^{N} \pi_n  e^{- (\mathbf{q}_T - \boldsymbol{\mu}^k_n)^{\top} \boldsymbol{\Sigma}^{k, -1}_n  (\mathbf{q}_T - \boldsymbol{\mu}^k_n)}} \boldsymbol{\mu}^v_n \\
    &= \sum^N_{n=0}\frac{c_n e^{-\frac{\beta}{2}||\mathbf{q}_T - \boldsymbol{\mu}^k_n||^2}}{\sum^J_{j=0} c_j e^{-\frac{\beta}{2} ||\mathbf{q}_T - \boldsymbol{\mu}^k_j||^2}} \boldsymbol{\mu}^v_n \\
    &= \sum^N_{n=0}\frac{c_n e^{\beta \mathbf{q}_T^{\top} \boldsymbol{\mu}^k_n}}{\sum^J_{j=0} c_j e^{\beta \mathbf{q}_T^{\top} \boldsymbol{\mu}^k_j}} \boldsymbol{\mu}^v_n\\
    &= \sum^N_{n=0}\frac{e^{\beta \mathbf{q}_T^{\top} \boldsymbol{\mu}^k_n + \text{log}(c_n)}}{\sum^J_{j=0} e^{\beta \mathbf{q}_T^{\top} \boldsymbol{\mu}^k_j + \text{log}(c_j)}} \boldsymbol{\mu}^v_n\\
    &= \texttt{softmax}(\beta \mathbf{q}_T D_k^{\top} + \text{log}(\mathbf{c}))D_v,
\end{align}
where $\mathbf{c}$ is the vector of counts. The first step expands the probability terms and sets $\boldsymbol{\mu}^{v|\mathbf{q}_t}_n = \boldsymbol{\mu}^{v}_n$, which is true by assumption 1. The second step also applies assumption 1 to covariance term, and replaces $\pi_n$ with $c_n$, which is true in the case where learning is done via hard cluster assignments, which is the implied by the previous result. The third step applies our second assumption, and the finaly step simply moves the count term, $c_n$, inside the exponential.

\newpage
\section{Appendix B: Methods}
\subsection{VQ-attention Implementation}\label{app:dict_prelim}
\begin{wrapfigure}{r}{0.6\textwidth}
\includegraphics[width=0.59\textwidth]{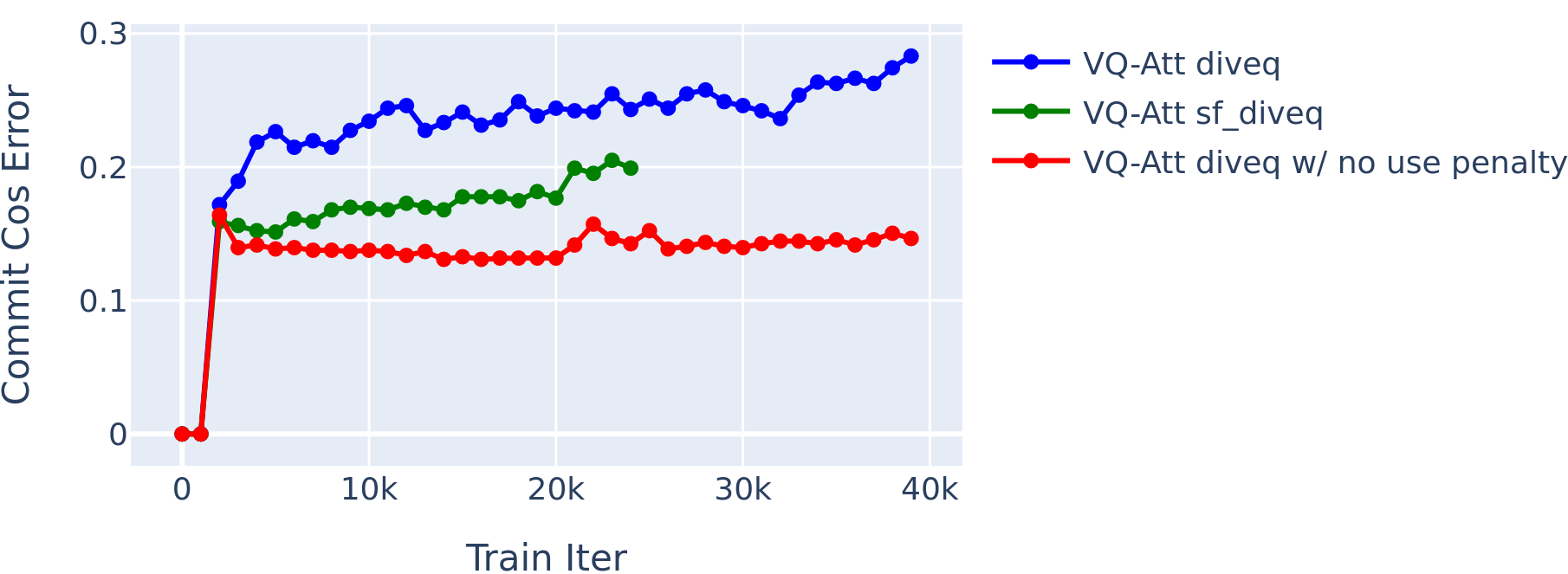}\caption{}\label{fig:app_commit_err}
\end{wrapfigure}

\textbf{Dictionary Training.} The original VQ-attention implementation \cite{lingleVQ} trained $D_k$ in pre-training using exponential moving average updates. It also used an auxiliary commitment loss similar to the original VQ-VAE \cite{van2017neural}. This approach is known to have several difficulties: it requires tuning numerous extra hyper-parameters, it suffers from dead neuron problems (some centroids stop getting updated), and dictionary updates are not computed during the backward pass but are instead computed outside the computation graph, making incorporation into parallelized training tricky. We use a recent SOTA method for training dictionaries in VQ-VAE layers, known as differentiable vector quantization (DiVeq) \cite{vali2025diveq}, that does not require auxiliary losses and computes dictionary updates through the backward pass. The same work \cite{vali2025diveq} adds a method to help prevent dead neurons, known as space filling DiVeq (SF-DiVeq). SF-DiVeq updates a weighted combination of centroids instead of single centroids, to increase the number of columns in the dictionary updated, thereby reducing chances of dead centroids. We found SF-DiVeq prevented dead centroids better than DiVeq, but was imperfect. To further improve, we simply add a penalty each training iteration to those centroids that have zero magnitude updates, eventually forcing dead centroids to be updated. See figure \ref{fig:app_commit_err}. Finally, we normalize centroids and keys to unit norm, which is known to further ease cluster and prevent negative effects of outliers \cite{kumar2010clustering}. Following the recommendations of \cite{vali2025diveq}, we first train the model for 4-9 thousand iterations without quantized keys, then begin quantizing keys and training the dictionary afterward. 

\textbf{Architecture.} The original implementation of VQ-attention uses an non-standard architecture in which a single head with small key dimension (i.e., 128) and very large value dimension (2D) are used. Learned position encoding are applied to the sliding window but NoPE is used outside the sliding window. Equation \ref{eq:vq_quad} shows the only mathematical difference between standard self-attention and VQ-attention are the quantized keys. In our tests, we aim to isolate the effects of quantizing keys from the other non-standard architecture choices of the original VQ transformer architecture (e.g., non-standard head number, non-standard head dimensions, learned position encodings, etc.). To do so, we use a VQ-attention layer that uses multiple heads and identically sized keys and values. All models use queries and keys that are unit norm, multiplied by learned, per head, scalar $\beta$. The only difference between a self-attention layer with NoPE and the VQ-attention layers we use, is the fact that VQ quantizes keys, while the baseline does not. 

\subsection{Sliding Window - NoPE Interleave Implementation}
We focus on hybrid architectures that interleave sliding window attention with RoPE and VQ, OVQ, and standard self attention with NoPE in an alternating pattern. We focus on this architecture in our experiments for two reasons. First, we find \texttt{sw-nope} was consistently our strongest baseline, and yielding also the best performance in OVQ architectures. Second, OVQ attention layers use position encodings in a different way than standard attention (see below). This means that OVQ-attention layers that use RoPE will not have two differences from standard attention layers with RoPE 1) quantized keys and 2) position encodings. We wanted to isolate the effects of quantizing keys in particular in our experiments, which we could do in the hybrid architectures that we tested. All models use queries and keys that are unit norm, multiplied by learned, per head, scalar $\beta$. Sliding window are size 128 on all models, unless specified otherwise. Head dimension is 128 in all models.

\subsection{OVQ-attention with NoPE Implementation}\label{app:ovq_implement} OVQ-attention use queries and keys that are unit norm, multiplied by learned, per head, scalar $\beta$. Our OVQ-attention layer uses NoPE. We use chunk size $L=128$ in all experiments. We sketch a naive implementation in pythonic pseudo-code below. 
\begin{lstlisting}
def ovq_attention(Q_chunks, K_chunks, V_chunks, dict_sizes):
    #Q_chunks: [Q_0, Q_1, ...Q_C]
    #K_chunks: [K_0, K_1, ...K_C]
    #V_chunks: [V_0, V_1, ...V_C]
    #dict_sizes: [N_L, N_{2L}, ..., N_{CL}] via equation 17
    
    #Initialize GMR components
    D_k = empty(B, H, dict_sizes[-1], d)
    D_v = empty(B, H, dict_sizes[-1], d)
    counts = empty(B, H, dict_sizes[-1], 1)
    ones = empty(B, H, L, 1)
    
    outs = []
    for c in range(C):
        # Calculate OVQ-attention using current chunk and dictionary, see equation 15
        d_sz = dict_sizes[c]
        attn_out = causal_attention(Q_chunks[c], 
                                    cat([D_k[:,:,0:d_sz,:], K_chunks[c]]), 
                                    cat([D_v[:,:,0:d_sz,:], V_chunks[c]]), 
                                    cat([counts[:,:,0:d_sz,:], ones]))
        
        outs.append(attn_out)
        
        # Update dictionary states for next chunk
        num_new = dict_sizes[c+1] - dict_sizes[c]
        update_dict(D_k, D_v, counts, K_chunks[c], V_chunks[c], num_new, d_sz)

    return outs
\end{lstlisting}
The dictionary update first gets nearest neighbor assignments for each key in the current chunk and the existing centroids in the dictionary. It then uses a series of gather and scatter add operations to updated the counts vector and to perform a mini-batch update on the dictionary centroids being updated. In preliminary tests, we found keeping the centroids normalized does not noticably effect performance, so we list that step as optional.
\begin{lstlisting}
def update_dict(D_k, D_v, counts, K_c, V_c, ones, num_new, d_sz):
    # Get nearest neighbor assignments for each key-value
    nn_assignments = get_nn_assignments(D_k[:,:,0:d_sz,:], 
                                        K_c, 
                                        num_new)

    #Update count vector
    counts.scatter_add(2, nn_assignments, ones)

    #Get lr for each centroid update
    lr = 1.0 / counts.gather(2, nn_assignments)

    #Get nearest neighbor centroid for each key value
    k_quant = torch.gather(D_k, 2, best_cluster)
    v_quant = torch.gather(D_v, 2, best_cluster)

    #Update dictionaries, equation 19
    D_k.scatter_add(2, best_cluster, -lr * (k_quant - K_c))
    D_v.scatter_add(2, best_cluster, -lr * (v_quant - V_c))

    #Normalization (optional)
    Dk = Dk.normalize(dim=-1)
    Dv = Dv.normalize(dim=-1)
\end{lstlisting}
Nearest neighbor assignments are obtained by getting the dot products between keys, $\mathbf{k}_t$, and centroids in $D_k$. Our theory suggests we should be comparing keys and values, $[\mathbf{k}_t, \mathbf{v}_t]$ to both dictionaries, $[D_k, D_v]$. However, we found that just using key dot products works well in practice and halves the compute cost.
\begin{lstlisting}
def get_nn_assignments(D_k, K_c, num_new, d_sz):
    #Current dictionary size
    N_c = D_k.size(2)
    
    # Get nearest neighbor assignments for each key-value
    sim = einsum('bhsd,bhnd->bhsn', K_c, D_k)
    best_sim, best_cluster = sim.max(dim=3)            

    # Lowest-similarity keys -> new centroids
    low_sim_idx = best_sim.topk(k=num_new, dim=-1, smallest=True).indices

    # Map those sequence positions -> their new centroid ids
    new_ids = arange(N_c, N_c + num_new) 
    new_ids = new_ids.view(1,1,-1).expand(B,H,-1) 

    # Best_cluster overwrite: scatter new centroid ids into those positions
    # Low_sim_idx gives sequence indices, so we scatter into dim=2 of best_cluster
    best_cluster.scatter(2, low_sim_idx, new_ids)
    
    return best_cluster
\end{lstlisting}

\subsection{OVQ-attention w/ RoPE Implementation}
OVQ-attention w/ RoPE is identical to the above except that 1) instead of concatenating just the current chunk, $K_c$, to the dictionary we concatenate the current and previous chunk, and 2) RoPE is applied to the current chunk and previous chunk. In particular, we set all centroids to have position 0, so they get the same rotation, then keys and queries within the current and previous window have position 1 through 257 (for chunk size 128). This is similar to the original method for VQ-attention \cite{lingleVQ}, which applied NoPE to the dictionary but learned additive position encodings to the current and previous chunk. Part of the motivation for using NoPE on dictionary elements is that keys and values will cluster better without position encodings (e.g., two identical token embeddings can yield keys that cluster well without position encodings but can cluster poorly when position encodings are applied).

\subsection{Long In-Context Recall Task}
\textbf{Basic ICR.} The context is filled with key-value pairs. Each key and each value are unique and consist of 8 tokens. A special token marks the assignment of one key to a value ($\rightarrow$ in the diagram below), and another special token marks the next key-value pair ('$|$' in the diagram below). At the end of the context, a query token marks the start of a query section, where copies of keys from the context are presented and the model must predict the correct value tokens. We use a vocab size of 10,000.
\begin{align}
&\textbf{Input } \ \text{ K1}\rightarrow \text{V1 } \ | \ \text{ K2}\rightarrow \text{V2 } \ | \ \text{ K3}\rightarrow \text{V3 } \ | \ \text{ K4}\rightarrow \text{V4 } \ \ | \ \text{ K5}\rightarrow \text{V5 } \ | \ \text{ K6}\rightarrow \text{V6 } \ |....| \ \text{ K6}\rightarrow \text{V6 } \ | \ \text{ K2}\rightarrow \text{V2}\\
&\textbf{Target}\ \ \phi \ \ \ \phi \ \ \ \phi \ \ \ \phi \ \ \ \phi \ \ \ \phi \ \ \ \phi \ \ \ \phi \ \ \ \phi \ \ \ \phi \ \ \ \phi \ \ \ \phi \ \ \ \phi \ \ \ \phi \ \ \ \phi \ \ \ \phi \ \ \ \phi \ \ \ \phi \ \ \ \phi \ \ \ \phi \ \ \ \phi \ \ \ \phi \ \ \ \phi \ \ \ \phi \ \ \ \phi \ \ \phi \ \ \text{V6} \ \ \phi \ \ \ \phi \ \ \ \phi \ \ \text{V2} \ \ \phi 
\end{align}
\textbf{Positional ICR} task has the same setup except in the context there are four copies of each key, and each copy is assigned a distinct value. In the query section four copies of one key are presented and the model must output all the values in the order they appear in the context. 
\begin{align}
&\textbf{Input } \ \text{ K1}\rightarrow \text{V1 } \ | \ \text{ K2}\rightarrow \text{V2 } \ | \ \text{ K3}\rightarrow \text{V3 } \ | \ \text{ K2}\rightarrow \text{V4 } \ \ | \ \text{ K3}\rightarrow \text{V5 } \ | \ \text{ K1}\rightarrow \text{V6 } \ |....| \ \text{ K2}\rightarrow \text{V2 } \ | \ \text{ K2}\rightarrow \text{V4}\\
&\textbf{Target}\ \ \phi \ \ \ \phi \ \ \ \phi \ \ \ \phi \ \ \ \phi \ \ \ \phi \ \ \ \phi \ \ \ \phi \ \ \ \phi \ \ \ \phi \ \ \ \phi \ \ \ \phi \ \ \ \phi \ \ \ \phi \ \ \ \phi \ \ \ \phi \ \ \ \phi \ \ \ \phi \ \ \ \phi \ \ \ \phi \ \ \ \phi \ \ \ \phi \ \ \ \phi \ \ \ \phi \ \ \ \phi \ \ \phi \ \ \text{V2} \ \ \phi \ \ \ \phi \ \ \ \phi \ \ \text{V4} \ \ \phi 
\end{align}

\begin{table}[h]
    \centering
    \begin{tabular}{lcccccccc} 
        \toprule
        \textbf{Model} & \textbf{Params} & \textbf{LR} & \textbf{Layers} & \textbf{Model Dim} & \textbf{Key Dim} & \textbf{Value Dim} & \textbf{Heads} & \textbf{MLP Size} \\ 
        \midrule
        \texttt{sw-nope} & 77M & $6e^{-4}$ & $8$ & $768$ & $128$ & $128$ & $6$ & $2304$\\
        \texttt{sw-vq 1k} & 80M & $4e^{-4}$ & $8$ & $768$ & $128$ & $128$ & $6$ & $2304$\\
        \texttt{sw-vq 2k} & 83M & $4e^{-4}$ & $8$ & $768$ & $128$ & $128$ & $6$ & $2304$\\
        \texttt{sw-vq 3k} & 86M & $4e^{-4}$ & $8$ & $768$ & $128$ & $128$ & $6$ & $2304$\\
        \texttt{sw-ovq} & 77M & $6e^{-4}$ & $8$ & $768$ & $128$ & $128$ & $6$ & $2304$\\
        \texttt{gdn} & 77M & $6.5e^{-4}$ & $8$ & $768$ & $128$ & $256$ & $3$ & $2304$\\
        \texttt{mesa-net} & 77M & $4e^{-4}$ & $8$ & $768$ & $128$ & $128$ & $6$ & $2304$\\
        \texttt{mamba2} & 77M & $4e^{-4}$ & $8$ & $768$ & $64$ & $64$ & $12$ & $2304$\\
        \bottomrule
    \end{tabular}
    \caption{Hyper-parameters used for Basic ICR and Positional ICR. Learning rates are found via a grid search. Cosine decay used with \texttt{min-lr=.00001}.}
\end{table}

\subsection{Long In-Context Learning Task} 
Our ICL task is builds on linear regression ICL tasks \cite{akyurek2022learning} to make the task require greater long-context ICL abilities. Like these previous works, in our version, the context consists of input-output pairs, where the output of each is generated by some linear function, $func_f(x)$. Each function performs three operations on the input tokens: it shuffles the tokens, it multiples by some integer between $0 < a < 6$, and adds an integer $0 < b < 6$. These operations can be expressed as a single linear regression operation $func_f(x) = b + a P \mathbf{x}$, where the input $\mathbf{x}$ is a column vector of positive integers and $P$ is a permutation matrix. We use a vocab size of 10,000. Input and output are 12 tokens each. 128 special tokens are used as function identifiers ($\rightarrow_{fn}$ in diagram below), and one token is used as a next example token ('$|$' in the diagram below). Examples are I.I.D. throughout the context.
\begin{align}
&\textbf{Input } \text{ I1}\rightarrow_{f4} \text{O1 } \ | \  \ \text{ I2}\rightarrow_{f1} \text{O2 } \ | \ \ \text{ I3}\rightarrow_{f3} \text{O3 } \ | \ \text{ I4}\rightarrow_{f4}  \text{O4 } \ | \ \text{ I5}\rightarrow_{f2} \text{O5 } \ | \  \ \text{ I6}\rightarrow_{f3} \text{O6 } ....\\
&\textbf{Target}\ \phi \ \ \ \text{O1} \ \ \ \phi \ \ \ \phi \ \ \ \phi \ \ \ \text{O2} \ \ \ \phi \ \ \ \phi \ \ \ \phi \ \ \ \text{O3} \ \ \ \phi \ \ \ \phi \ \ \ \phi \ \ \ \text{O4} \ \ \ \phi \ \ \ \phi \ \ \ \phi \ \ \  \text{O5} \ \ \ \phi \ \ \ \phi \ \ \ \phi \ \ \ \text{O6}\ \ \phi ....
\end{align}

\begin{table}[h]
    \centering
    \begin{tabular}{lcccccccc} 
        \toprule
        \textbf{Model} & \textbf{Params} & \textbf{LR} & \textbf{Layers} & \textbf{Model Dim} & \textbf{Key Dim} & \textbf{Value Dim} & \textbf{Heads} & \textbf{MLP Size} \\ 
        \midrule
        \texttt{sw-nope} & 77M & $4e^{-4}$ & $8$ & $768$ & $128$ & $128$ & $6$ & $2304$\\
        \texttt{sw-vq 1k} & 80M & $4e^{-4}$ & $8$ & $768$ & $128$ & $128$ & $6$ & $2304$\\
        \texttt{sw-vq 2k} & 83M & $4e^{-4}$ & $8$ & $768$ & $128$ & $128$ & $6$ & $2304$\\
        \texttt{sw-vq 3k} & 86M & $4e^{-4}$ & $8$ & $768$ & $128$ & $128$ & $6$ & $2304$\\
        \texttt{sw-ovq} & 77M & $6e^{-4}$ & $8$ & $768$ & $128$ & $128$ & $6$ & $2304$\\
        \texttt{gdn} & 77M & $6.5e^{-4}$ & $8$ & $768$ & $128$ & $256$ & $3$ & $2304$\\
        \texttt{mesa-net} & 77M & $2.5e^{-4}$ & $8$ & $768$ & $128$ & $128$ & $6$ & $2304$\\
        \texttt{mamba2} & 77M & $8e^{-4}$ & $8$ & $768$ & $64$ & $64$ & $12$ & $2304$\\
        \bottomrule
    \end{tabular}
    \caption{Hyper-parameters used for ICL tasks. Learning rates found through grid search. Cosine decay used with \texttt{min-lr=.00001}.}
\end{table}

\subsection{PG19 Language Modeling}
We filter the PG19 dataset \cite{rae2016scaling} to only include sequence 16k or longer, ensuring the model always sees full coherent sequences during both training and testing. We train models at sequence length 10k and test at 16k. Models are tested every 1000 iterations. The best test score achieved is the one shown in figure \ref{fig:pg19}. In the figure, we average over bins of length 500. Models have model dimension of 1024. Attention layers use eight heads sized 128 each. Following default hyper-parameters, the GDN layers use four heads but with double the size values, i.e., keys sized 128 and values sized 256 doubling each state size.
\begin{table*}[h]
    \centering
    \begin{tabular}{lcccccccc} 
        \toprule
        \textbf{Model} & \textbf{Params} & \textbf{LR} & \textbf{Layers} & \textbf{Model Dim} & \textbf{Key Dim} & \textbf{Value Dim} & \textbf{Heads} & \textbf{MLP Size} \\ 
        \midrule
        Attn models & 257M & $8e^{-4}$ & $17$ & $1024$ & $128$ & $128$ & $8$ & $2304$\\
        GDN models & 257M & $8e^{-4}$ & $17$ & $1024$ & $128$ & $[128, 256]$ & $[8, 4]$ & $2304$\\
        \bottomrule
    \end{tabular}
    \caption{PG19 Hyper-parameters.  Cosine decay used with \texttt{min-lr=.00001}. In \texttt{gdn models} sliding window layers use 128 size value dim and 8 heads, while \texttt{gdn} layer use 256 value dim and 4 heads.}
\end{table*}

\subsection{Short Context Benchmarks}
\begin{table}[h]
    \centering
    \begin{tabular}{lcccccccc} 
        \toprule
        \textbf{Model} & \textbf{Params} & \textbf{LR} & \textbf{Layers} & \textbf{Model Dim} & \textbf{Key Dim} & \textbf{Value Dim} & \textbf{Heads} & \textbf{MLP Size} \\ 
        \midrule
        \texttt{sw-nope} & 480M & $8e^{-4}$ & $21$ & $1280$ & $128$ & $128$ & $10$ & $3200$\\
        \texttt{sw-ovq} & 480M & $8e^{-4}$ & $21$ & $1280$ & $128$ & $128$ & $10$ & $3200$\\
        \bottomrule
    \end{tabular}
    \caption{Hyper-parameters usd for short context benchmarks.Cosine decay used with \texttt{min-lr=.00002}.}
\end{table}
Models are trained on 50B tokens from fineweb-edu \cite{penedo2024fineweb} tokenized using the Mistral tokenizer. Train sequence length is 1536 with batch size 256. 

\newpage

\section{Appendix C: Further Results}\label{app:further_results}

\begin{wrapfigure}{r}{0.5\textwidth}
\includegraphics[width=0.48\textwidth]{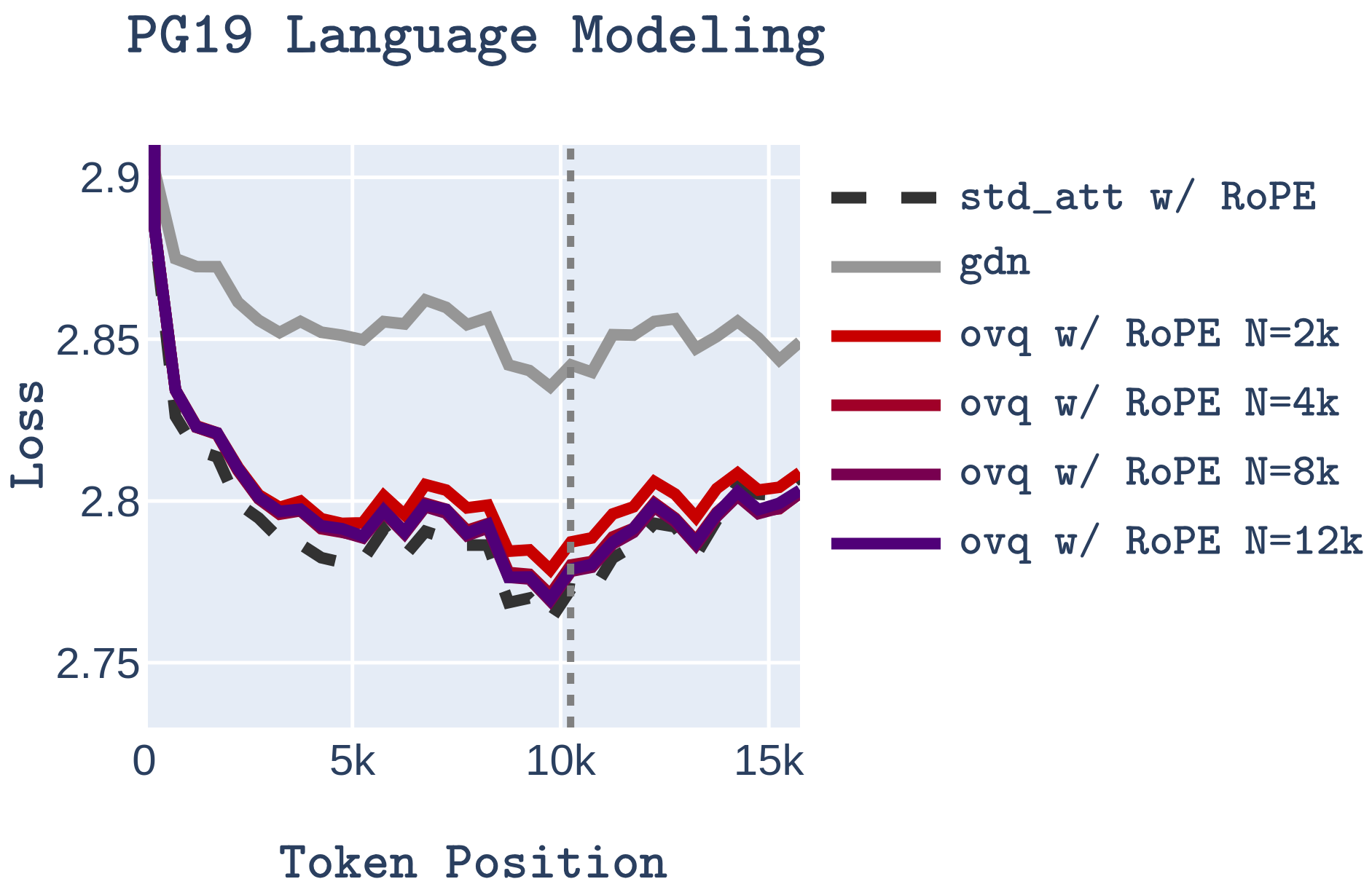}\caption{}
\end{wrapfigure}
\textbf{OVQ with RoPE on PG19.} We focused our experiments on hybrid architectures that interleaved OVQ layers that used NoPE with SWA and GDN layers. This allowed us to isolate the effects of architectural changes, i.e., the only difference between the interleaved models that use VQ and OVQ attention with NoPE versus those that use full attention with NoPE are the quantized keys (see CITE for explanation). We also test models that use only OVQ-attention layers with RoPE position encodings. Like the original VQ-attention \cite{lingleVQ}, we apply position encodings to the current and previous block and NoPE to the dictionary (see CITE for explanation). We use RoPE position encodings. We compare to a pure GDN architecture and a pure self-attention architecture that uses RoPE. We see cross-entropy loss is drastically better for OVQ-attention model than for GDN and is very similar between OVQ-attention and standard attention model up to 16k tokens, even when OVQ-attention uses N as small as 4k.\\ \\ \\

\begin{wrapfigure}{r}{0.5\textwidth}
\includegraphics[width=0.48\textwidth]{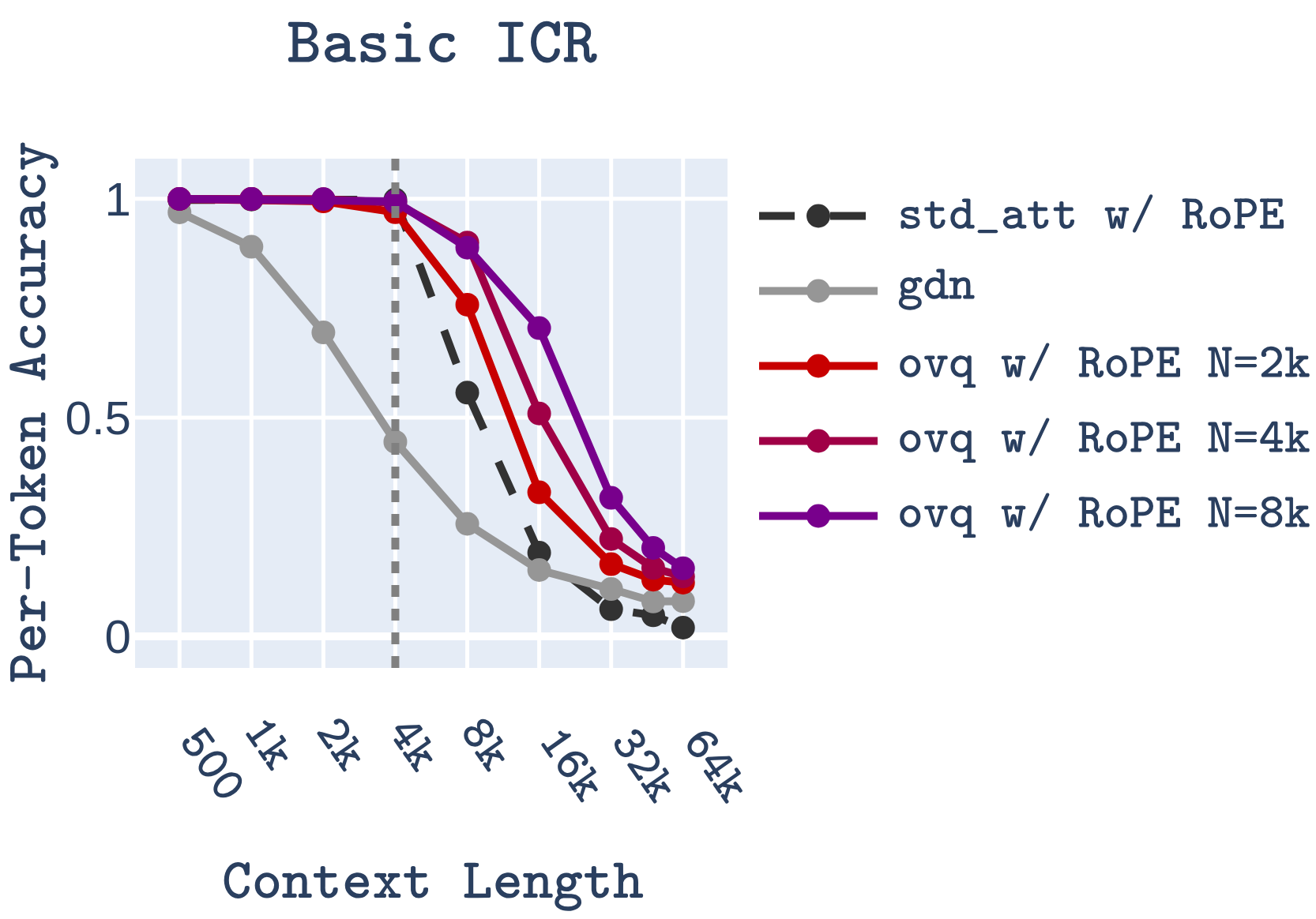}\caption{}
\end{wrapfigure}
\textbf{OVQ with RoPE on Basic Recall.} We also test OVQ with RoPE on the basic recall task. We observe that OVQ models are able to solve basic recall up to the train length. The model that uses standard attention layers with RoPE length generalize very poorly. OVQ models with RoPE also struggles to length generalize, though its performance degradation is not as extreme. This is likely due to the fact the OVQ models only apply RoPE to a fixed size window of recent tokens and NoPE for token elements further in the past represented in the dictionary. \\ \\ \\ \\ \\ \\ \\ \\

\begin{wrapfigure}{r}{0.5\textwidth}
\includegraphics[width=0.48\textwidth]{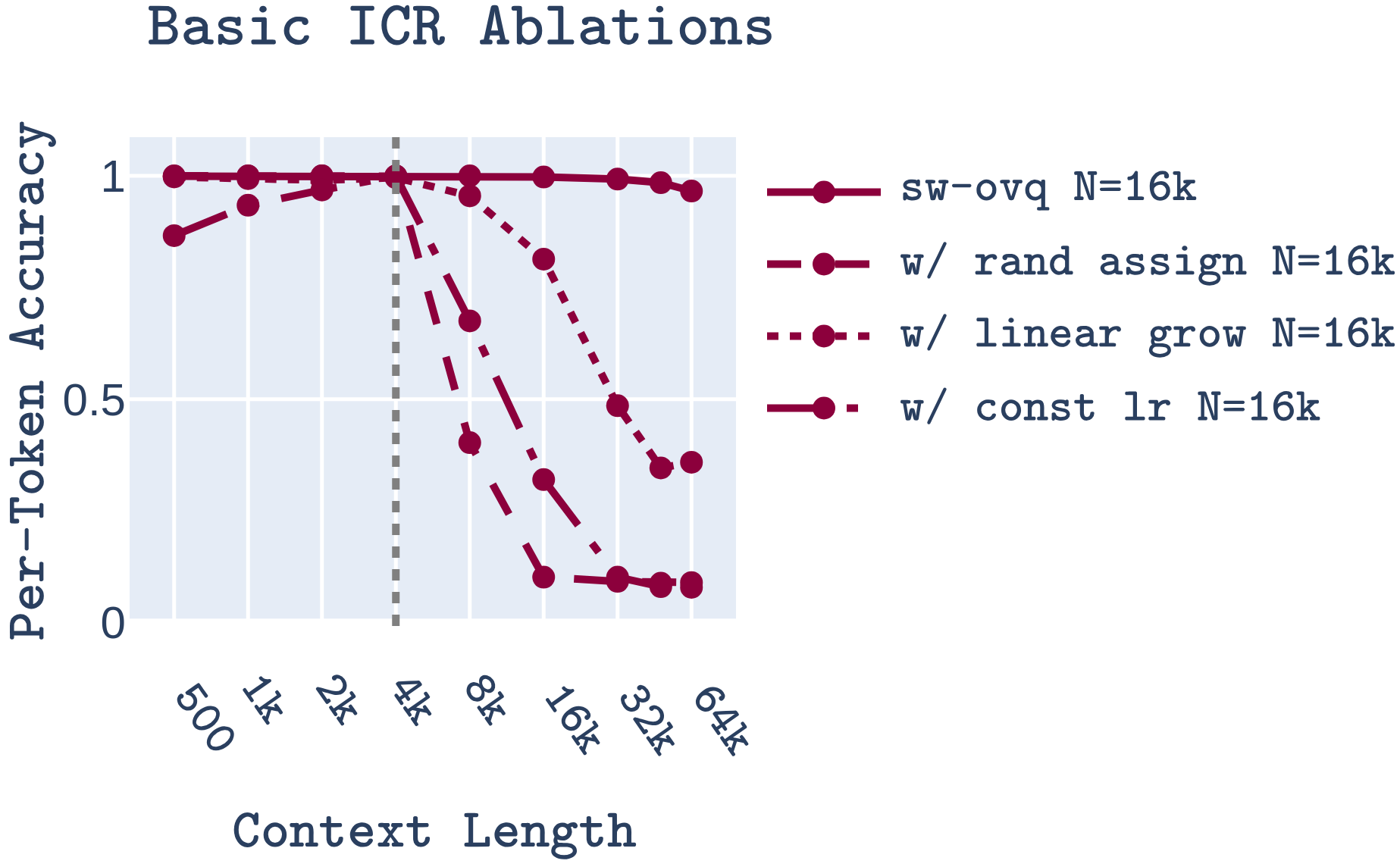}\caption{}
\end{wrapfigure}
\textbf{Ablations on Basic ICR.} We test three ablations on OVQ-attention using the basic ICR task. First, instead of assigning setting new centroids based on our spread maximizing scheme, we set the new centroids equal to a random sample of key values within the current chunk (\texttt{rand assign}). Second, instead of using our plateauing dictionary growth scheme where centroids are added quickly early in the sequence then slow later on, we add new centroids linearly, where every chunk in the sequence adds the same number of new centroids (\texttt{linear grow}). Finally, instead of using the adapting learning rates for updating dictionary centroids, we use a constant learning rate of .025 (\texttt{const lr}). Using a constant learning rate is equivalent to doing gradient descent, instead of our Newton's method, on the k-means loss. We can see that each one of these ablations significantly worsens the models ability to recall beyond 4k context. \\ \\

\begin{wrapfigure}{r}{0.5\textwidth}
\includegraphics[width=0.48\textwidth]{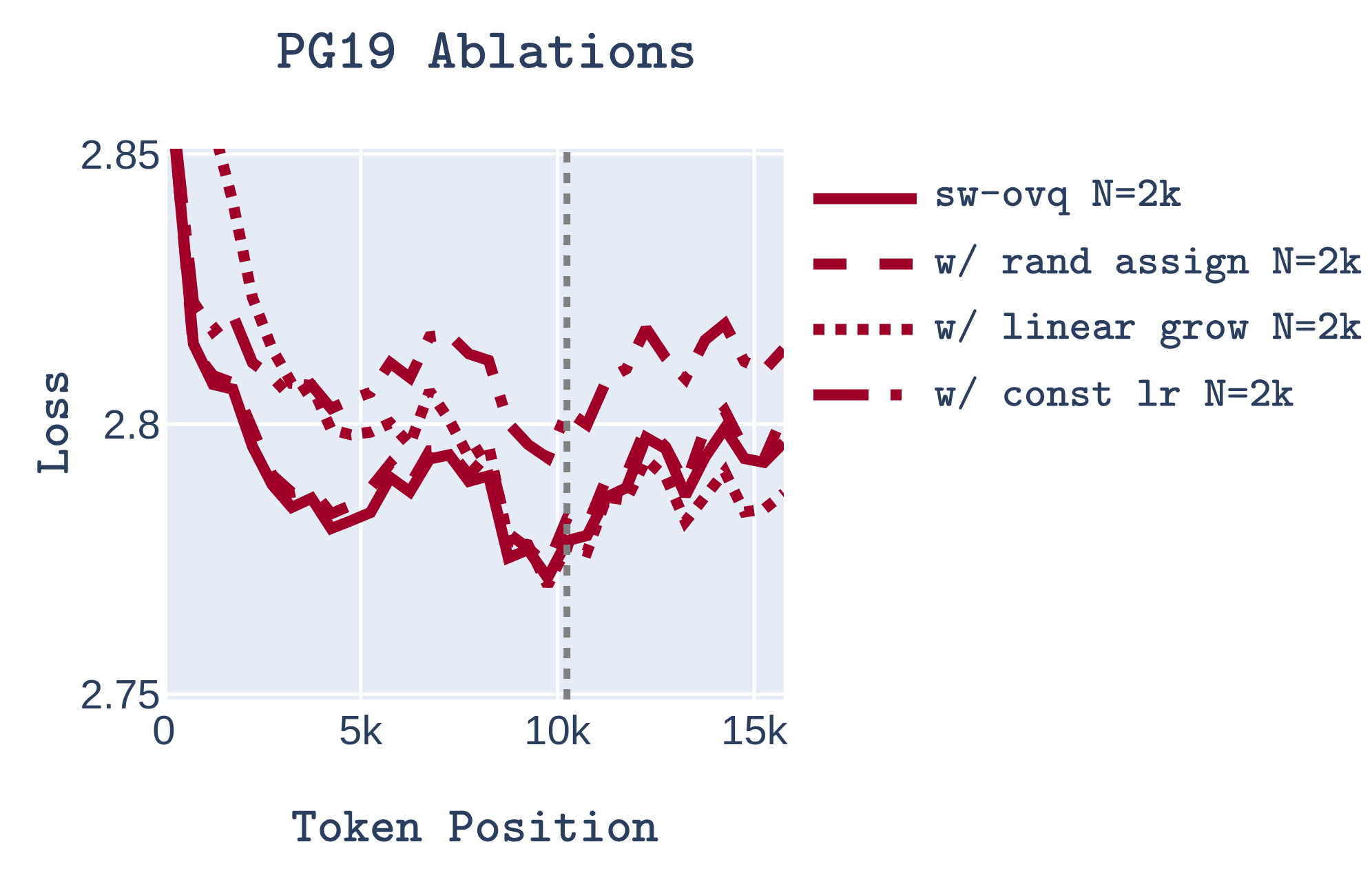}\caption{}
\end{wrapfigure}
\textbf{Ablations on PG19.} We test the same three ablations on PG19 task. We use the same set up as above, with 10k train length and 16k test length. Model size 257M are used. Each model is trained to convergence. The linear dictionary growth version does noticeably worse than our OVQ attention at short lengths. It does catch up, however, and matches, even slightly surpasses at very long lengths, likely because it grows much slower than our method early in the sequence, and faster later in the sequence. The constant learning rate method method is also consistently worse across the sequence than our method for initializing the dictionary in a way that maximizes spread. Interestingly, in this task, the version of OVQ with random assignments is about the same as the original, which uses adaptive learning rates. \\ \\

\begin{wrapfigure}{r}{0.5\textwidth}
\includegraphics[width=0.48\textwidth]{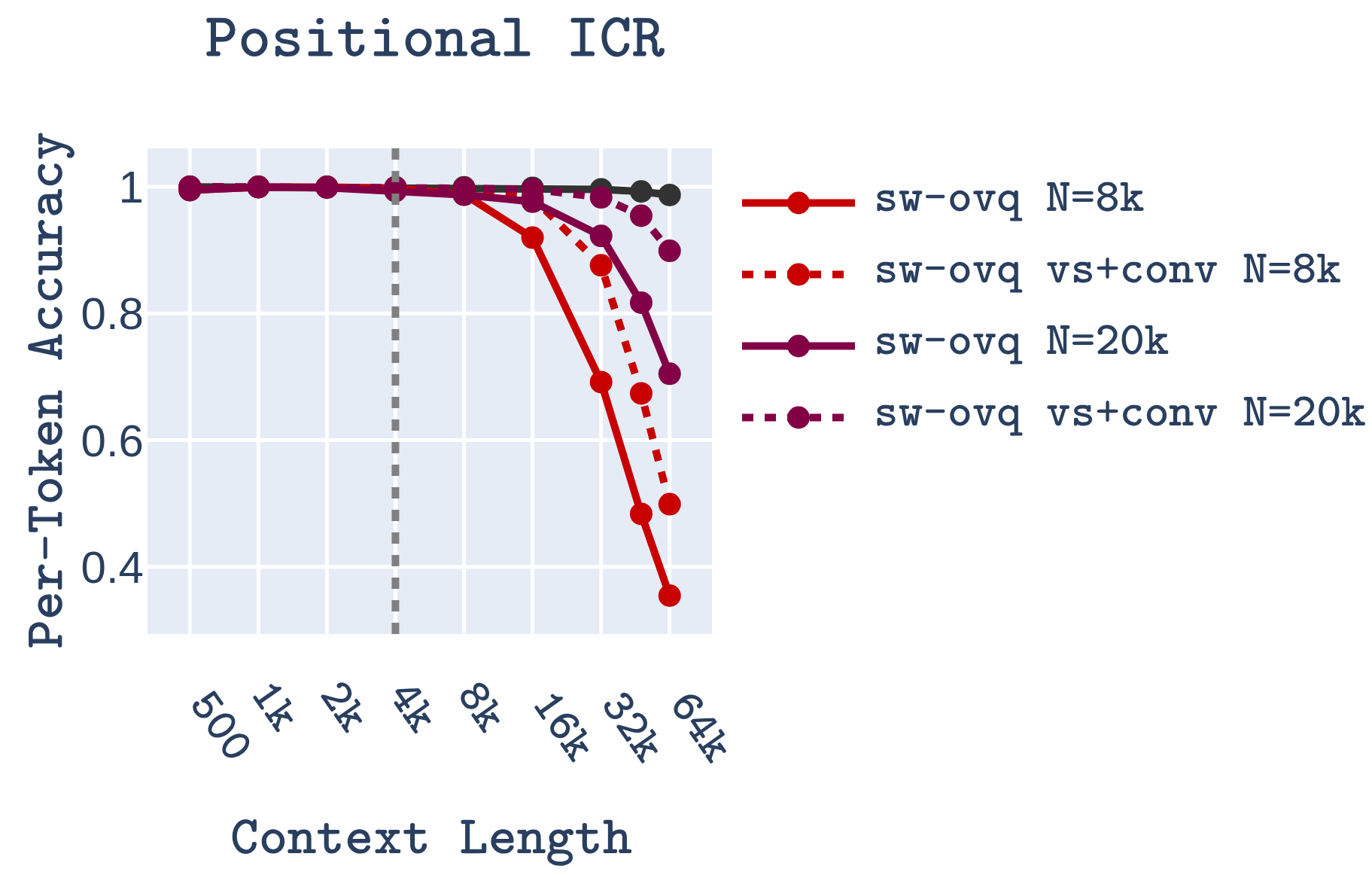}
  \caption{}\label{fig:vshift_conv}
\end{wrapfigure}
\textbf{Preliminary Testing of V-shifting and Convolutions.} Inspired by linear attention models like RWKV \cite{peng2023rwkv} and self attention variants (e.g., \cite{zhang2024memory, figliolia2025compressed}) we test the effects of applying convolutions to query and keys, and shifting values forward in time, so $\mathbf{k}_t$ is associated with $\mathbf{v}_{t+1}$. Following \cite{zhang2024memory}, each $\mathbf{k}_t$ is associated with a weighted average of the current and future value: $\mathbf{v}_{t+\frac{1}{2}} = \sigma(\alpha) \mathbf{v}_t + (1-\sigma(\alpha)) \mathbf{v}_{t+1}$, where $\sigma$ is a sigmoid function and $\alpha$ is a learned scalar. Keys and values are then both shifted to prevent the model from breaking causality. We test on the positional ICR task, the task OVQ-attention struggles on the most at long context lengths. Results are in figure \ref{fig:vshift_conv}. Interestingly, we find applying convolutions and v-shifting significantly improves length extrapolation at long test lengths. These results provide preliminary evidence that the ability of OVQ-attention to perform and generalize to long test lengths might be further improved via simple architecture tweaks. Future work can explore alterations like this in more detail. \\

\textbf{Testing Dictionary Training Methods.} We test several SOTA methods to pre-train the dictionaries in VQ-attention layers: DiVeq and  SF-DiVeQ \cite{vali2025diveq}. We also test DiVeq with a "no use penalty", where we add a scalar penalty to the similarity values of each centroid. This pentalty increases by .0025 with every training iteration that the centroid does not get updated. This increases the chances dead neurons/centroids get a nearest neighbor assignment and an update. Commitment error is the average cosine similarity between keys and their nearest neighbor centroid, averaged over batch, head, sequence, and layers.

\newpage
\section{Appendix D: Flops Analysis}\label{app:flops}

\begin{figure}[h]
  \begin{center}
\includegraphics[width=0.6\textwidth]{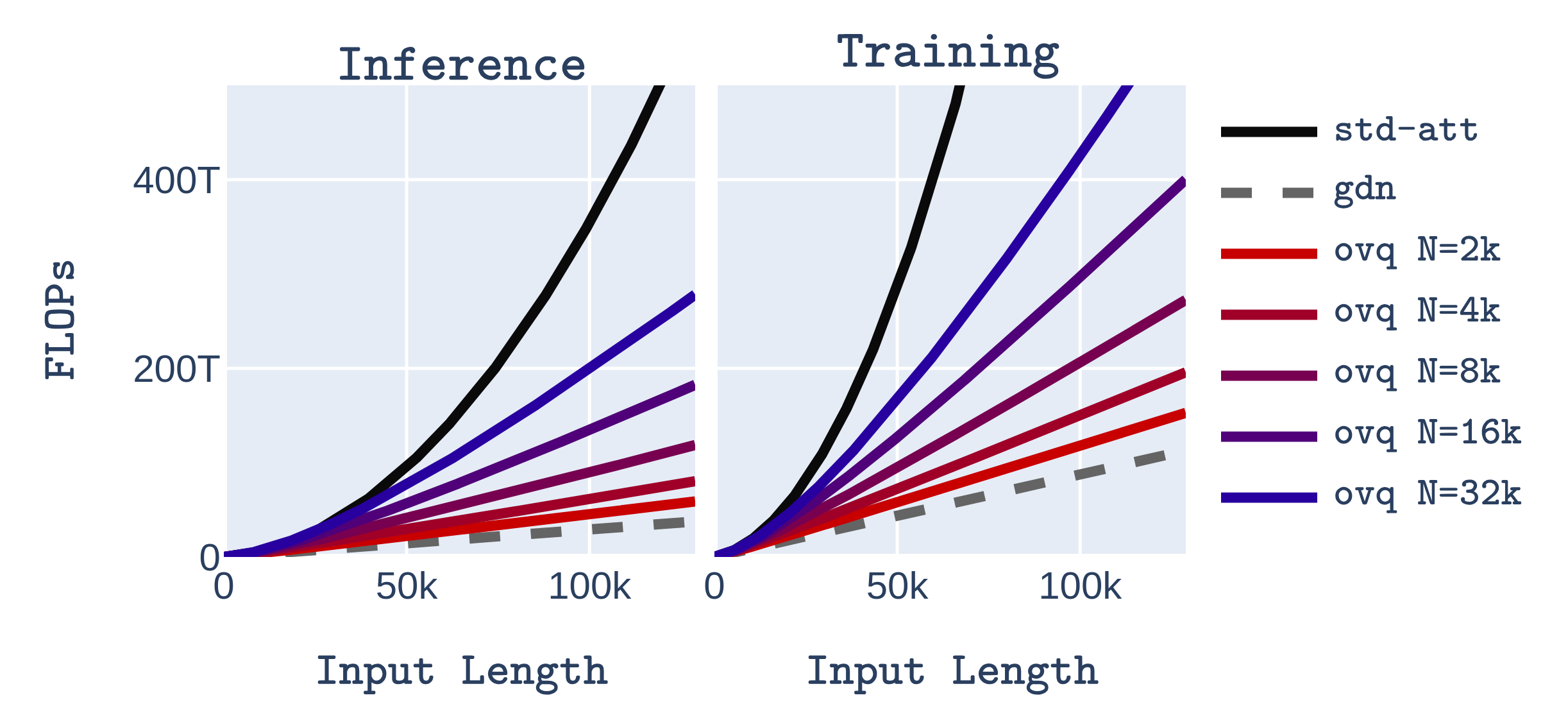}
\caption{Theoretical flops at different context lengths.}\label{fig:ovq_learn}
  \end{center}
\end{figure}

\begin{figure}[h]
  \begin{center}
\includegraphics[width=0.6\textwidth]{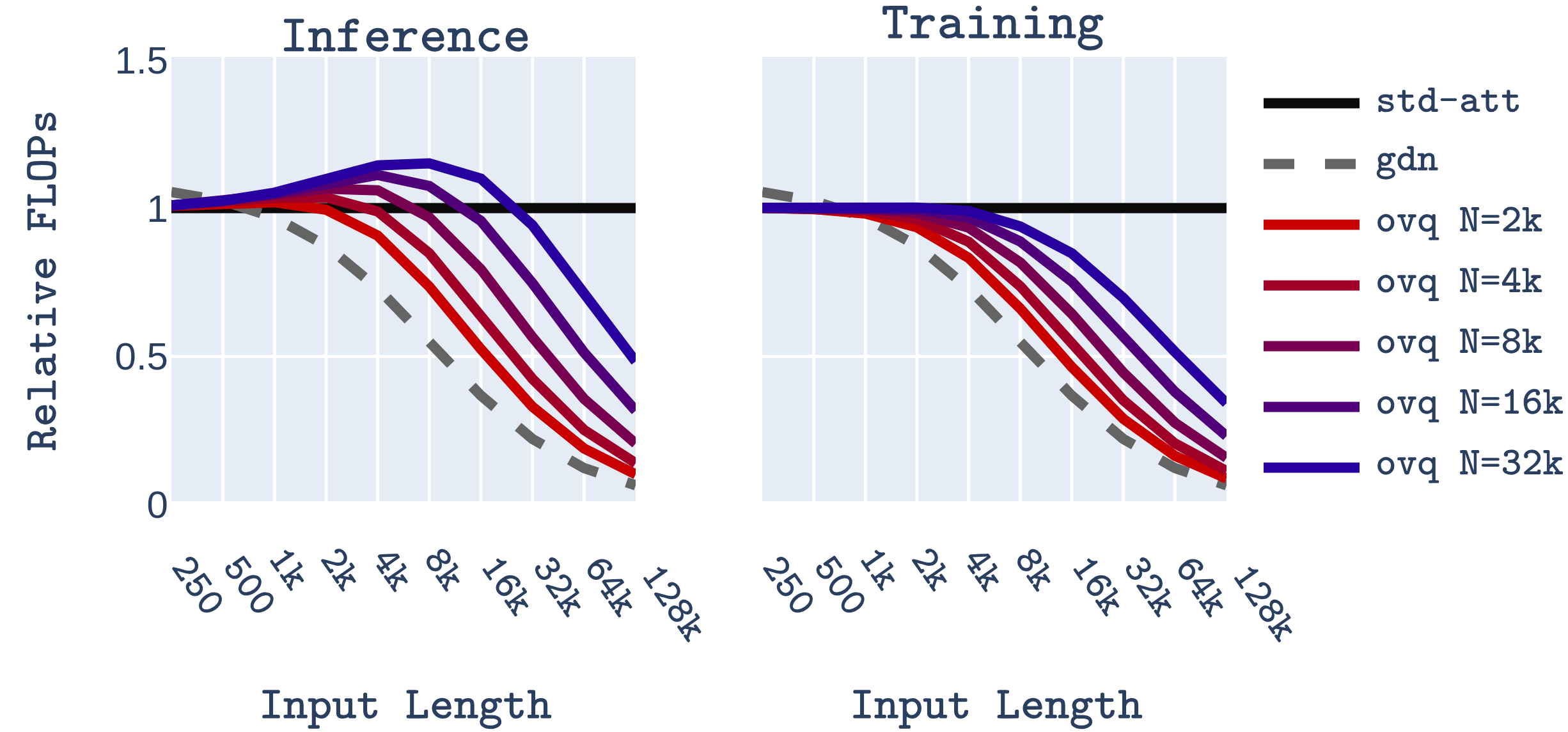}
\caption{Flops ratio, where standard attention is used as the reference/denominator.}\label{fig:ovq_learn}
  \end{center}
\end{figure}

\begin{table}[h]
\centering
\begin{tabular}{l}
\hline
\textbf{Notation}  \\\hline
\textbf{Symbol} \ \ \ \ \ \ \textbf{Meaning}\\\hline
$B$   Batch size\\
$H$   Number of heads\\
$T$   Sequence length\\
$d$   Head dimension\\
$L$   Chunk Size\\
$C$   Number of Chunks \\
$N_c$ Dictionary size at chunk $c$\\
\hline
\end{tabular}
\caption{Notation}
\end{table}

In the above plots, is a comparison of the flops used in OVQ-attention, self-attention, and gated delta net. Below are details on how these are computed. We analyse the sequence mixing operations specifically, i.e., given Q, K, V what are the flops required to produce the output of the sequence mixing operation?

\begin{table}[h]
\centering
\begin{tabular}{l|l|l}
\hline
&\textbf{Causal Self-Attention FLOPs}  \\\hline
Operation & Inference Flops & Train Flops \\\hline
$S = QK^{\top}$ & $2 B H T^2 d / 2$ & $6 B H T^2 d$ \\
$A = \texttt{softmax}(S)$ & $B H T^2$(Neg.) & $3B H T^2 / 2$(Neg.)\\
$A V$ & $BHT^2d$ & $3BHT^2d / 2$\\
\textbf{Attn Total} & $2BHT^2d / 2$ & $6BHT^2d / 2$ \\
\hline
\end{tabular}
\caption{Flops required for causal self-attention.}
\end{table}

\begin{table}[h]
\centering
\begin{tabular}{l|l|l}
\hline
&\textbf{OVQ Attention Per Chunk FLOPs}  \\\hline
Operation & Inference Flops & Train Flops \\\hline
$Q_c D_{k,c}^{\top}$ & $2 B H L N_c d$ & $6 B H L N_c d$ \\
$Q_c K_c^{\top} \odot M$ & $B H L^2 d$ & $3B H L^2 d$ \\
$\texttt{softmax}$ & $BHL(N_c+L)$ (Neg.) & $3BHL(N_c+L)$ (Neg.)\\
$Attn \text{ x } D_{v,c}$ & $2B H L N_c d$ & $6B H L N_c d$\\ 
$Attn \text{ x } V_c$ & $B H L^2 d$ & $3B H L^2 d$ \\ 
\textbf{Attn Total} & $BHLd (4 N_c + 2L)$ & $BHLd (12 N_c + 6L)$\\ \hline
$S_c = K_c D_{k,c}^{\top}$ & $2B H L N_c d$ & $0$\\
$m_c = \texttt{argmax}(S_c)$ & $BHLN_c$ (Neg.) & $0$\\
$\texttt{top-k}(m_c, n_{new,c})$ & $BHL \text{log}_2(n_{new,c})$ (Neg.) & 0\\
$\texttt{scatter-add}(C, \texttt{ones}, m_c)$ & $BHL$ (Neg.) & 0 \\
$\texttt{scatter-add}(\gamma_c (\hat{K}_c - K_c))$ & $3BHLN_c$ (Neg.) & $9BHLN_c$ (Neg.)\\
$\texttt{scatter-add}(\gamma_c (\hat{V}_c - V_c))$ & $3BHLN_c$ (Neg.) & $9BHLN_c$ (Neg.)\\
\textbf{Update Total} & $2BHLN_cd$ & (Neg.)\\ \hline
$\textbf{Combined Total}$ & $\mathbf{BHLd (6 N_c + 2L)}$ & $\mathbf{BHLd (12 N_c + 6L)}$\\ \hline
\end{tabular}
\caption{\label{tab:widgets} Flops per chunk, $c$. In the table 'Neg.' means negligible compared to the most costly operation.}
\end{table}

\begin{equation}
\text{OVQ Total Inference FLOPs} = \sum_c^C BHLd (6 N_c + 2L)
\end{equation}

\begin{equation}
\text{OVQ Total Training FLOPs} = \sum_c^C BHLd (12 N_c + 6L)
\end{equation}

We use the same flops calculations for gated delta net as previous works using the same set up we have here (e.g., \cite{lufkin2026hybrid, yang2024gated}), which comes out to
\begin{equation}
\text{Inference GDN Flops} = 6BTHd^2 + BTH(6d^2 + 2L5d + L^2/3).
\end{equation}
\begin{equation}
\text{Training GDN Flops} = 18BTHd^2 + 3BTH(6d^2 + 2L5d + L^2/3).
\end{equation}

\end{document}